\documentclass[10pt,journal,compsoc]{IEEEtran}
\ifCLASSOPTIONcompsoc
  \usepackage[nocompress]{cite}
  \usepackage{graphicx}
  \usepackage{amsmath}
  \usepackage{amssymb}
  \usepackage{bm}
  \usepackage{color}
  \usepackage{array}
  \usepackage{multirow}
  \usepackage{ragged2e}

\else
  \usepackage{cite}
\fi
\ifCLASSINFOpdf
\else
\fi
\begin{document}
%
\title{Cross Euclidean-to-Riemannian Metric Learning with Application to Face Recognition from Video}
%
%
%
%

\author{Zhiwu~Huang,~\IEEEmembership{Member,~IEEE,}
        Ruiping~Wang, ~\IEEEmembership{Member,~IEEE,}
        Shiguang~Shan,~\IEEEmembership{Senior Member,~IEEE,}
        Luc~Van Gool, ~\IEEEmembership{Member,~IEEE}
        and ~Xilin~Chen, ~\IEEEmembership{Fellow,~IEEE}
\IEEEcompsocitemizethanks{\IEEEcompsocthanksitem Zhiwu Huang was with the Key Lab of Intelligent Information Processing of Chinese Academy of Sciences (CAS), Institute of Computing Technology (ICT), Beijing, 100190, China. He is now with the Computer Vision Laboratory, Swiss Federal Institute of Technology (ETH), Zurich, 8092, Switzerland. E-mails: zhiwu.huang@vision.ee.ethz.ch.}
\IEEEcompsocitemizethanks{\IEEEcompsocthanksitem Ruiping Wang, Shiguang Shan and Xilin Chen are with the Key Lab of Intelligent Information Processing of Chinese Academy of Sciences (CAS), Institute of Computing Technology (ICT), Beijing, 100190, China. E-mails: \{wangruiping, sgshan, xlchen\}@ict.ac.cn. (Corresponding author: Shiguang Shan.)}
\IEEEcompsocitemizethanks{\IEEEcompsocthanksitem Luc Van Gool is with the Computer Vision Laboratory, Swiss Federal Institute of Technology (ETH), Zurich, 8092, Switzerland. E-mails: vangool@vision.ee.ethz.ch.}}
\IEEEtitleabstractindextext{%

\begin{abstract}
\justifying{Riemannian manifolds have been widely employed for video representations in visual classification tasks including video-based face recognition. The success mainly derives from learning a discriminant Riemannian metric which encodes the non-linear geometry of the underlying Riemannian manifolds. In this paper, we propose a novel metric learning framework to learn a distance metric across a Euclidean space and a Riemannian manifold to fuse the average appearance and pattern variation of faces within one video. The proposed metric learning framework can handle three typical tasks of video-based face recognition: Video-to-Still, Still-to-Video and Video-to-Video settings. To accomplish this new framework, by exploiting typical Riemannian geometries for kernel embedding, we map the source Euclidean space and Riemannian manifold into a common Euclidean subspace, each through a corresponding high-dimensional Reproducing Kernel Hilbert Space (RKHS). With this mapping, the problem of learning a cross-view metric between the two source heterogeneous spaces can be expressed as learning a single-view Euclidean distance metric in the target common Euclidean space. By learning information on heterogeneous data with the shared label, the discriminant metric in the common space improves face recognition from videos. Extensive experiments on four challenging video face databases demonstrate that the proposed framework has a clear advantage over the state-of-the-art methods in the three classical video-based face recognition tasks.}

\end{abstract}

}
\maketitle

\IEEEdisplaynontitleabstractindextext

%
\IEEEpeerreviewmaketitle

\IEEEraisesectionheading{\section{Introduction}\label{intro}}

%
%
%
%

\IEEEPARstart{D}{ue} to robustness against varying imaging conditions, Riemannian manifolds have proven powerful representations for video sequences in many branches of computer vision. Two of the most popular Riemannian structures are the manifold of linear subspaces (i.e., Grassmann manifold) and the manifold of Symmetric Positive Definite (SPD) matrices. From a different perspective, these Riemannian representations can be related to modeling a video with a multivariate Gaussian distribution, characterized by its mean and covariance matrix. In the case of the Grassmann manifold, the distances between subspaces can be reduced to distances between multivariate Gaussian distributions by treating linear subspaces as the flattened limit of a zero-mean, homogeneous factor analyzer distribution \cite{hamm2008extended}. In the case of the SPD manifold, a sequence of video frames is represented as the covariance matrix of the image features of frames \cite{wang2012covariance, vemulapalli2013kernel}, which therefore essentially encodes a zero-mean Gaussian distribution of the image features. In \cite{huang2015face, wang2015discriminant}, each video is modeled as a Gaussian distribution with non-zero mean and covariance matrix, which can be combined to construct an SPD matrix, and thus also resides on a specific SPD manifold \cite{amari2000methods, Lovric2000jvma}.


The success of Riemannian representations in visual recognition is mainly due to the learning of more discriminant metrics, which encode Riemannian geometry of the underlying manifolds. For example, by exploiting the geometry of the Grassmann manifolds, \cite{hamm2008gda} proposed Grassmann kernel functions to extend the existing Kernel Linear Discriminant Analysis \cite{fukunaga2013introduction} to learn a metric between Grassmannian representations. In \cite{huang2015projection}, a new method is presented to learn a Riemannian metric on a Grassmann manifold by performing a Riemannian geometry-aware dimensionality reduction from the original Grassmann manifold to a lower-dimensional, more discriminative Grassmann manifold where more favorable classification can be achieved. To learn a discriminant metric for an SPD representation, \cite{wang2012covariance} derived a kernel function that explicitly maps the SPD representations from the SPD manifold to a Euclidean space where a traditional metric learning method such as Partial Least Squares \cite{wold1985partial} can be applied. In \cite{harandi2014manifold}, an approach is proposed to search for a projection that yields a low-dimensional SPD manifold with maximum discriminative power, encoded via an affinity-weighted similarity measure based on Riemannian metrics on the SPD manifold.

\begin{figure}[t]
\begin{center}
\includegraphics[width=1\linewidth]{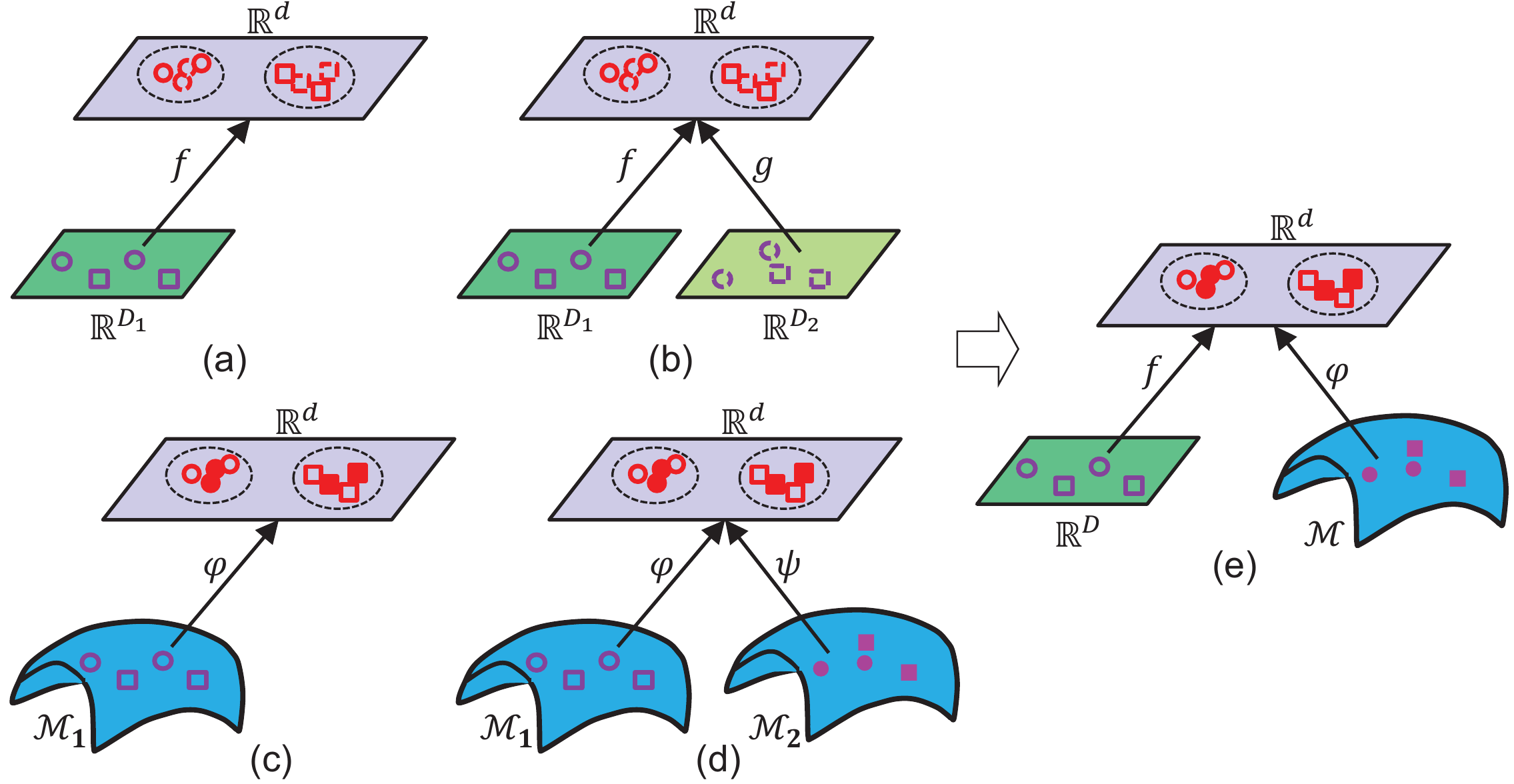}
\end{center}
\caption{Conceptual illustration of traditional Euclidean metric learning (a), Euclidean-to-Euclidean metric learning (b), Riemannian metric learning (c), Riemannian-to-Riemannian metric learning (d) and the proposed cross Euclidean-to-Riemannian metric learning (e). $\mathbb{R}^{D_1}/\mathbb{R}^{D_2}$, $\mathcal{M}/\mathcal{M}_1/\mathcal{M}_2$ and $\mathbb{R}^d$ indicate a Euclidean \mbox{space}, a Riemannian manifold and a common subspace, respectively. $f/g,\varphi/\psi$ denote linear and non-linear transformations, and \mbox{different} shapes (i.e., circles and rectangles) represent classes.}
\label{Fig1}
\end{figure}

In this paper, we focus on studying the application of Riemannian metric learning to the problem of video-based face recognition, which identifies a subject with his/her face video sequences. Generally speaking, there exist three typical tasks of video-based face recognition, i.e., Video-to-Still (V2S), Still-to-Video (S2V) and Video-to-Video (V2V) face identification/verification. Specifically, the task of V2S face recognition matches a query video sequence against still face images, which are typically taken in a controlled setting. This task commonly occurs in watch list screening systems. In contrast, in the task of S2V face recognition, a still face image is queried against a database of video sequences, which can be applied to locate a person of interest by searching his/her identity in the stored surveillance videos. The third task, i.e., V2V face recognition, looks for the same face in input video sequences among a set of target video sequences. For example, one could track a person by matching his/her video sequences recorded somewhere against the surveillance videos taken elsewhere.



To handle the tasks of video-based face recognition, state-of-the-art deep feature learning methods \cite{taigman2014deepface,schroff2015facenet,parkhi2015deep,sun2015deepid3,wen2016discriminative} typically adopt mean pooling strategy to fuse deep features from single frames within one face video. However, in addition to the first-order mean pooling, as studied in many works such as \cite{hamm2009extended,lu2013image,liu2014combining,huang2015face,wang2015discriminant}, the pattern variation (i.e., second-order pooling) of videos provides yet an important complementary cue for video-based face recognition. With this motivation in mind, we propose a new metric learning scheme across a Euclidean space and a Riemannian manifold to match/fuse appearance mean and patter variation (i.e., first- and second-order poolings) for still images and video sequences. In particular, the learning scheme employs either raw or even deeply learned feature vectors (i.e., Euclidean data) from still facial images, while representing the faces within one video sequence with both their appearance mean (i.e., Euclidean data) and pattern variation models that are typically treated as Riemannian data. Benefited from the new architecture, the three typical video-based face recognition tasks can be uniformly tackled. Compared with the previous version \cite{huang2014lerm} that can only handle V2S/S2V face recognition with pattern variation modeling on videos, this paper mainly makes two technical improvements:
\begin{itemize}

  \item  To improve the V2S/S2V face recognition task, the new framework represents each video simultaneously with appearance mean and pattern variation models, and derives a more generalized cross Euclidean and Riemannian metric learning scheme to match still images and video sequences.
  \item The original framework is also extended to handle the task of V2V face recognition. To this end, the objective function of the framework is adapted to fuse Euclidean data (i.e., appearance mean) and Riemannian data (i.e., pattern variation) of videos in a unified framework.

\end{itemize}



The key challenge of learning Euclidean-to-Riemannian metric learning is the essentially  heterogeneous properties of the  processed underlying data spaces, i.e., Euclidean spaces and Riemannian manifolds, which respect totally different geometrical structures and thus are equipped with different metrics, i.e. Euclidean distance and Riemannian metric respectively. As a result, applying most of traditional metric learning methods in Fig.\ref{Fig1}(a), (b), (c) and (d) will totally break down in the context of learning a metric across a Euclidean space and a Riemannian manifold. For example, Euclidean-to-Euclidean metric learning Fig.\ref{Fig1}(b) merely learns a discriminative distance metric between two Euclidean spaces with different data domain settings, while Riemannian-Riemannian metric learning Fig.\ref{Fig1}(c) only explores a discriminative function across two homogeneous Riemannian manifolds. Hence, in the metric learning theory, this work mainly brings the following three innovations:
\begin{itemize}

  \item  As depicted in Fig.\ref{Fig1}(e), a novel heterogeneous metric learning framework is developed to match/fuse Euclidean and Riemannian representations by designing a new objective function well performing across Euclidean-Riemannian spaces. To the best of our knowledge, it is one of the first attempts to learn the metric across Euclidean and Riemannian spaces.
  \item The proposed metric learning scheme can accommodate a group of typical non-Euclidean (Riemannian) representations widely used in vision problems, e.g., linear subspaces, affine subspaces and SPD matrices. Thus, it is a general metric learning framework to study the problem of fusing/matching hybrid Euclidean and Riemannian data.

\end{itemize}

\section{Related Work} \label{rel}

In this section we review relevant Euclidean metric learning and Riemannian metric learning methods. In addition, we also introduce existing applications of Riemannian metric learning to the problem of video-based face recognition.

\subsection{Euclidean Metric Learning}

In conventional techniques to learn a metric in a Euclidean space, the learned distance metric is usually defined as a Mahalanobis distance, which is the squared Euclidean distance after applying the learned linear transformation(s) to the original Euclidean space(s). According to the number of the source Euclidean spaces, traditional metric learning methods can be categorized into Euclidean metric learning and Euclidean-to-Euclidean metric learning.

As shown in Fig.\ref{Fig1}(a), the Euclidean metric learning methods \cite{goldberger2004neighbourhood,davis2007information,sugiyama2007dimensionality,weinberger2009distance,zhu2013p2s} intend to learn a metric or a transformation $f$ of object features from a source Euclidean \mbox{space} $\mathbb{R}^{D}$ to a target Euclidean space $\mathbb{R}^{d}$. For example, \cite{davis2007information} introduced an information-theoretic formulation to learn a metric in one source Euclidean space. In \cite{weinberger2009distance}, a metric learning method was proposed to learn a transformation from one Euclidean space to a new one for the K-nearest neighbor algorithm by pulling neighboring objects of the same class closer together and pushing others further apart. In \cite{zhu2013p2s}, an approach was presented to learn a distance metric between (realistic and virtual) data in a \mbox{single} Euclidean space for the definition of a more appropriate point-to-set distance in the application of point-to-set classification.

In contrast, as depicted in Fig.\ref{Fig1}(b), the Euclidean-to-Euclidean metric learning methods \cite{hardoon2004canonical, bronstein2010data, sharma2011pls, quadrianto2011learning, mcfee2011learning, sharma2012gma, zhai2013heterogeneous} are designed to learn a cross-view metric or multiple \mbox{transformations} $f,g$ mapping object features from multiple source Euclidean \mbox{spaces}, say $\mathbb{R}^{D_1}$, $\mathbb{R}^{D_2}$, to a target common subspace $\mathbb{R}^{d}$. For instance, \cite{quadrianto2011learning} proposed a metric learning method to seek multiple projections under a neighborhood \mbox{preserving} constraint for multi-view data in multiple Euclidean spaces. In \cite{mcfee2011learning}, a multiple kernel/metric learning technique was applied to integrate different object features from multiple Euclidean spaces into a \mbox{unified} Euclidean space. In \cite{zhai2013heterogeneous}, a metric learning method was developed to learn two projections from two different Euclidean spaces to a common subspace by integrating the structure of cross-view data into a joint graph regularization.

\subsection{Riemannian Metric Learning}

Riemannian metric learning pursues discriminant functions on Riemannian manifold(s) in order to classify the Riemannian representations more effectively. In general, existing Riemannian metric learning Fig.\ref{Fig1} (c) and Riemannian-to-Riemannian metric learning Fig.\ref{Fig1} (d) adopt one of the following three typical schemes to achieve a more desirable Riemannian metric on/across Riemannian manifold(s).

The first Riemannian metric learning scheme \cite{tuzel2008pedestrian, tosato2010region, turaga2011statistical, carreira2012sem, sanin2013spatio, xu2014discriminative} typically first flattens the underlying Riemannian manifold via tangent space approximation, and then learns a discriminant metric in the resulting tangent (Euclidean) space by employing traditional metric learning methods. However, the map between the manifold and the tangent space is locally diffeomorphic, which inevitably distorts the original Riemannian geometry. To address this problem, LogitBoost on SPD manifolds \cite{tuzel2008pedestrian} was introduced, by pooling the resulting classifiers in multiple approximated tangent spaces on the calculated Karcher mean on Riemannian manifolds. Similarly, a weighted Riemannian locality preserving projection is exploited by \cite{sanin2013spatio} during boosting for classification on Riemannian manifolds.

Another family of Riemannian metric learning methods \cite{hamm2008gda, hamm2008extended, harandi2011ggda, wang2012covariance, vemulapalli2013kernel, li2013log, minh2014nips, jayasumana2015kernel, huang2015face} derives Riemannian metric based kernel functions to embed the Riemannian manifolds into a high-dimensional Reproducing Kernel Hilbert space (RKHS). As an RKHS respects Euclidean geometry, this learning scheme enables the traditional kernel-based metric learning methods to work in the resulting RKHS. For example, in \cite{hamm2008gda, hamm2008extended, harandi2011ggda, vemulapalli2013kernel}, the projection metric based kernel and its extensions were introduced to map the underlying Grassmann manifold to an RKHS, where kernel learning algorithms developed in vector spaces can be extended to their counterparts. To learn discriminant data on the SPD manifolds, \cite{wang2012covariance, vemulapalli2013kernel,li2013log, minh2014nips, jayasumana2015kernel, huang2015face} exploited some well-studied Riemannian metrics such as the Log-Euclidean metric \cite{arsigny2007led}, to derive positive definite kernels on manifolds that permit to embed a given manifold with a corresponding metric into a high-dimensional RKHS.

The last kind of Riemannian metric learning \cite{jung2012analysis, harandi2014manifold, huang2015projection, huang2015leml} learns the metric by mapping the original Riemannian manifold to another one equipped with the same Riemannian geometry. For instance, in \cite{harandi2014manifold}, a metric learning algorithm was introduced to map a high-dimensional SPD manifold into a lower-dimensional, more discriminant one. This work proposed a graph embedding formalism with an affinity matrix that encodes intra-class and inter-class distances based on affine-invariant Riemannian metrics \cite{pennect2006aid, sra2012new} on the SPD manifold. Analogously, on the Grassmann manifold, \cite{huang2015projection} proposed a new Riemannian metric learning to learn a Mahalanobis-like matrix that can be decomposed into a manifold-to-manifold transformation for geometry-aware dimensionality reduction.

In contrast to Riemannian metric learning performed on a single Riemannian manifold, Riemannian-to-Riemannian metric learning \cite{jayasumana2013combining, liu2014combining} typically learns multiple Riemannian metrics across different types of Riemannian manifolds by employing the second Riemannian metric learning scheme mentioned above. For example, in \cite{jayasumana2013combining}, multiple Riemannian manifolds were first mapped into multiple RKHSs, and a feature combining and selection method based on a traditional Multiple Kernel Learning technique was then introduced to optimally combine the multiple transformed data lying in the resulting RKHSs. Similarly, \cite{liu2014combining} adopted multiple traditional metric learning methods to fuse the classification scores on multiple Riemannian representations by employing Riemannian metric based kernels on their underlying Riemannian manifolds.

\subsection{Riemannian Metric Learning Applied to Video-based Face Recognition}

State-of-the-art methods \cite{hamm2008gda, hamm2008extended, hu2011sparse, turaga2011statistical, wang2012covariance, lu2013image,  harandi2014manifold, huang2015face, wang2015discriminant, huang2015leml} typically model each video sequence of faces with a variation model (e.g., linear subspace, affine subspace and SPD matrices) and learn a discriminant Riemannian metric on the underlying Riemannian manifold for robust video-based face recognition. For example, \cite{hamm2008gda} represented each set of video frames by a linear subspace of their image features. By exploiting the geometry of the underlying Grassmann manifold of linear subspaces, they extended the Kernel Linear Discriminant Analysis method to learn discriminative linear subspaces. As studied in \cite{hamm2008extended}, image sets are more robustly modeled by affine subspaces, each of which is obtained by adding an offset (i.e, the data mean) to one linear subspace. Analogously to \cite{hamm2008gda}, an affine Grassmann manifold and its Riemannian geometry were exploited by \cite{hamm2008extended} for affine subspace discriminant learning. In \cite{wang2012covariance}, each video is modeled as a covariance matrix, which is then treated as an SPD matrix residing on the SPD manifold. To learn discriminative SPD matrices, they applied traditional discriminant analysis methods such as Partial Least Squares on the manifold of SPD matrices. Accordingly, the success of these methods mainly derives from the effective video modeling with Riemannian representations and discriminant metric learning on such Riemannian data.


\section{Cross Euclidean-to-Riemannian Metric Learning} \label{cerml}

In this section, we first formulate the new problem of Cross Euclidean-to-Riemannian Metric Learning (CERML), and then present its objective function. In the end, we develop an optimization algorithm to solve the objective function.

\begin{figure}[t]
\begin{center}
\includegraphics[width=0.75\linewidth]{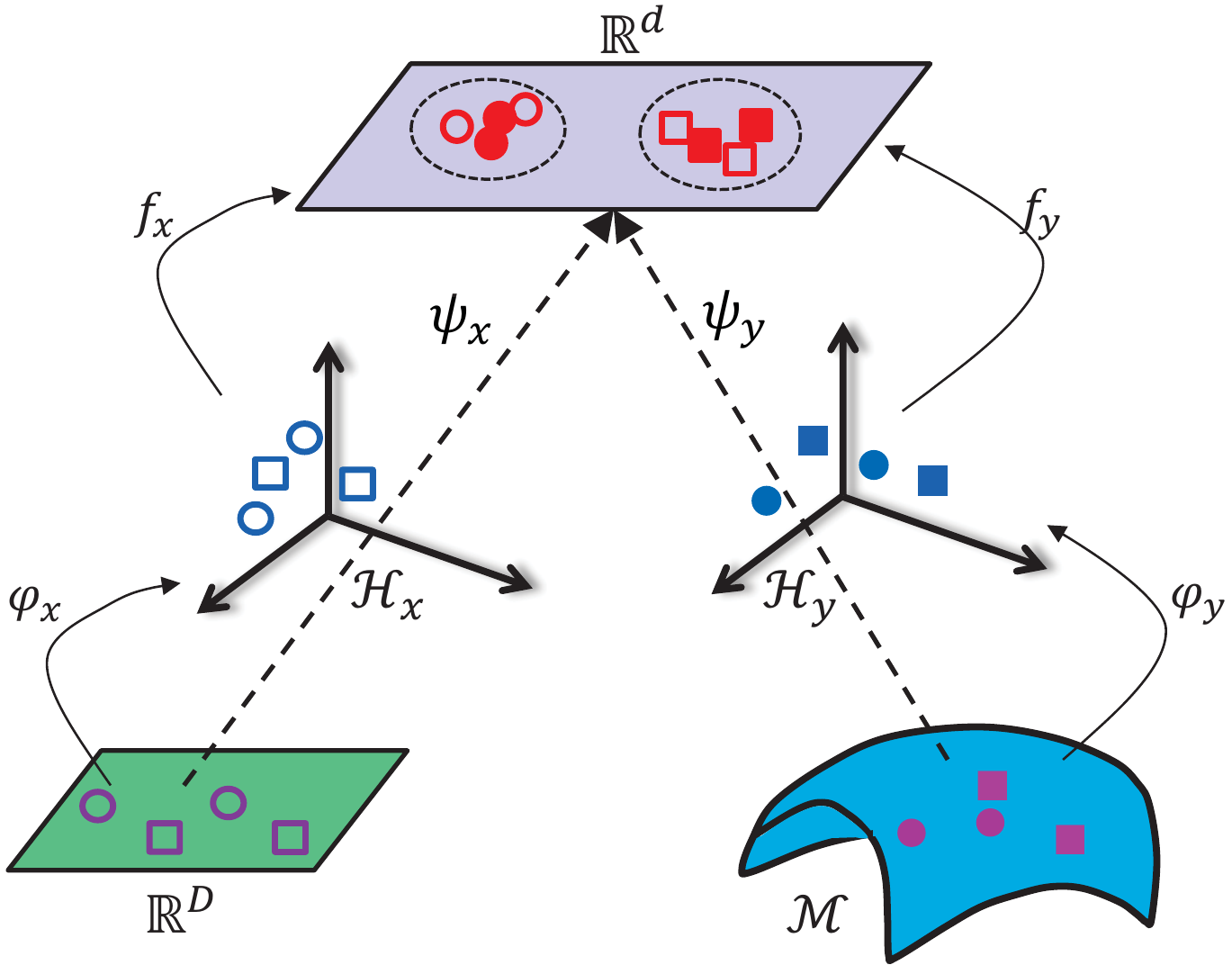}
\end{center}
\caption{Overview of the proposed Cross Euclidean-to-Riemannian Metric Learning (CERML) framework. $\mathbb{R}^{D}$, $\mathcal{M}$, $\mathcal{H}_{x}/\mathcal{H}_{{y}}$, $\mathbb{R}^{d}$ \mbox{represent} a Euclidean space, a Riemannian manifold, a Hilbert space and a common subspace respectively. $f_{x}/f_{{y}}$, $\psi_{x}/\psi_{{y}}$ denote linear and nonlinear transformation functions, and different shapes represent classes.}
\label{Fig2}
\end{figure}

\subsection{Problem Formulation} \label{form}

Let $\bm{X}=\{\bm{x}_1,\bm{x}_2,\ldots,\bm{x}_m\}, \bm{x}_i \in \mathbb{R}^{D}$ be a set of Euclidean data with class labels $\{l^x_1,l^x_2,\ldots,l^x_m\}$ and $\bm{Y}=\{\bm{y}_1,\bm{y}_2,\ldots,\bm{y}_n\}, \bm{y}_j \in \mathcal{M}$ be a set of Riemannian representations with class labels $\{l^{y}_1,l^{y}_2,\ldots,l^{y}_n\}$, where $\bm{y}_j$ come as a certain type of Riemannian representations such as linear subspaces, affine subspaces, or SPD matrices.

Given one pair of a Euclidean point $\bm{x}_i$ and a Riemannian point $\bm{y}_j$, we use $d(\bm{x}_i, \bm{y}_j)$ to represent their distance. To achieve an appropriate distance metric between them for better discrimination, we propose to learn \mbox{two} transformation functions $\psi_x$ and $\psi_y$, which respectively map the Euclidean points and Riemannian points to a common Euclidean subspace. In the common subspace, the learned distance metric between the involved pair of heterogeneous data can be reduced to the classical Euclidean distance as:
\begin{equation}
d(\bm{x}_i, \bm{y}_j) = \sqrt{(\psi_x(\bm{x}_i)-\psi_y(\bm{y}_j))^T(\psi_x(\bm{x}_i)-\psi_y(\bm{y}_j))}.
\label{Eq1}
\end{equation}

However, as the source Euclidean space $\mathbb{R}^{D}$ and Riemannian manifold $\mathcal{M}$ differ too much in terms of geometrical data structure, it is difficult to employ linear transformations to map them into the common target Euclidean subspace $\mathbb{R}^{d}$. This motivates us to first transform the Riemannian manifold to a flat Euclidean space so that the heterogeneity between this flattened space and the source Euclidean space reduces. To this end, there exist two strategies: one is tangent space embedding, and the other is Reproducing Kernel Hilbert Space (RKHS) embedding. The first strategy pursues an appropriate tangent space to approximate the local geometry of the Riemannian manifold. In contrast, the second strategy exploits typical Riemannian metrics based kernel functions to encode Riemannian geometry of the underlying manifold. As evidenced by the theory of kernel methods in Euclidean spaces, compared with the tangent space embedding scheme, the RKHS embedding yields much richer high-dimensional representations of the original data, making visual classification tasks easier.

With this idea in mind, and as shown in Fig.\ref{Fig2}, the proposed framework of Cross Euclidean-to-Riemannian Metric Learning (CERML) first derives the kernel functions based on typical Euclidean and Riemannian metrics to define the inner product of the implicit non-linear transformations $\varphi_x$ and $\varphi_y$, which respectively map the Euclidean space $\mathbb{R}^{D}$ and the Riemannian manifold $\mathcal{M}$ into two RKHSs $\mathcal{H}_{x}$, $\mathcal{H}_{y}$. After the kernel space embedding, two mappings $f_x, f_y$ are learned from the two RKHSs to the target common subspace. Thus, the final goal of this new framework is to employ the two mappings $\psi_x$ and $\psi_y$ to transform the original Euclidean data and Riemannian data into the common subspace $\mathbb{R}^{d}$, where the distance metric between each pair of Euclidean data point and Riemannian data point is reduced to the classical Euclidean distance defined in Eq. \ref{Eq1}. In particular, the two linear projections can be represented as $f_x(\bm{x}_i)=\bm{V}^T_x\bm{x}_i$, $f_y(\bm{y}_j)=\bm{V}^T_y\bm{y}_j$, where $\bm{V}^T_x, \bm{V}^T_y$ are two linear projection matrices. Inspired by the classical kernel techniques, we employ the corresponding kernel functions to derive the inner products of these two non-linear transformations as $\langle \varphi_x(\bm{x}_i),\varphi_x(\bm{x}_j)\rangle=\bm{K}_x(\bm{x}_i,\bm{x}_j), \langle \varphi_y(\bm{y}_i),\varphi_y(\bm{y}_j)\rangle=\bm{K}_y(\bm{y}_i,\bm{y}_j)$, where $\bm{K}_x,\bm{K}_y$ are the kernel matrices involved. By parameterizing the inner products in the two RKHSs, the formulations of the two final mapping functions $\psi_x$ and $\psi_y$ can be achieved by $\psi_x(\bm{x}_i)=\bm{W}^T_x \bm{K}_{x_{.i}}$, $\psi_y(\bm{y}_j)=\bm{W}^T_y \bm{K}_{y_{.j}}$, where $\bm{K}_{x_{.i}}, \bm{K}_{y_{.i}}$ are respectively the $i$-th columns of the kernel matrices $\bm{K}_x, \bm{K}_y$. Accordingly, the distance metric Eq.\ref{Eq1} between a pair of a Euclidean point $\bm{x}$ and a Riemannian representation $\bm{y}$ can be further formulated as:
\begin{equation}
d(\bm{x}_i, \bm{y}_j) = \sqrt{(\bm{W}^T_x \bm{K}_{x_{.i}}-\bm{W}^T_y \bm{K}_{y_{.j}})^T(\bm{W}^T_x \bm{K}_{x_{.i}}-\bm{W}^T_y \bm{K}_{y_{.j}})}.
\label{Eq2}
\end{equation}
Additionally, according to the above mapping mode, the distance metric between each pair of transformed homogeneous data points in the common Euclidean subspace can also be achieved as:
\begin{equation}
d_x(\bm{x}_i, \bm{x}_j) = \sqrt{(\bm{W}^T_x \bm{K}_{x_{.i}}-\bm{W}^T_x \bm{K}_{x_{.j}})^T(\bm{W}^T_x \bm{K}_{x_{.i}}-\bm{W}^T_x \bm{K}_{x_{.j}})},
\label{Eq3}
\end{equation}
\begin{equation}
d_y(\bm{y}_i, \bm{y}_j) = \sqrt{(\bm{W}^T_y \bm{K}_{y_{.i}}-\bm{W}^T_y \bm{K}_{y_{.j}})^T(\bm{W}^T_y \bm{K}_{y_{.i}}-\bm{W}^T_y \bm{K}_{y_{.j}})},
\label{Eq4}
\end{equation}
where the specific forms of $\bm{K}_{x}$ and $\bm{K}_{y}$ will be presented in the following.

Now, we need to define the kernel functions $\bm{K}_x(\bm{x}_i,\bm{x}_j), \bm{K}_y(\bm{y}_i,\bm{y}_j)$ for Euclidean data and Riemannian representations. For the Euclidean data, without loss of generality, we exploit the Radial Basis Function (RBF) kernel, which is one of the most popular positive definite kernels. Formally, given a pair of data point $\bm{x}_i,\bm{x}_j$ in Euclidean space, the kernel function is defined as:
\begin{equation}
\bm{K}_x(\bm{x}_i,\bm{x}_j) = exp(-\|\bm{x}_i-\bm{x}_j\|^2/2\sigma_x^2),
\label{K_Eq5}
\end{equation}
which actually employs the Euclidean distance between two Euclidean points $\bm{x}_i$ and $\bm{x}_j$.

As for the Riemannian representations, since they are commonly defined on a specific type of Riemannian manifold that respects a non-Euclidean geometry \cite{hamm2008gda, hamm2008extended, wang2012covariance}, the above kernel function formulation will fail. So, it has to be generalized to Riemannian manifolds. For this purpose, given two elements $\bm{y}_i,\bm{y}_j$ on a Riemannian manifold, we formally define a generalized kernel function for them as:
\begin{equation}
\bm{K}_y(\bm{y}_i,\bm{y}_j) = exp(-d^2(\bm{y}_i,\bm{y}_j)/2\sigma_y^2).
\label{K_Eq6}
\end{equation}

The kernel function performed on Riemannian representations actually takes the form of a Gaussian function (note that we also study the linear kernel case in the supplementary material). The most important component in such a kernel function is $d(\bm{y}_i,\bm{y}_j)$, which defines the distance between one pair of Riemannian points on the underlying Riemannian manifold. Next, this distance is discussed for three typical Riemannian representations, i.e., Grassmannian data, affine Grassmannian data and SPD data.

\setlength{\parskip}{0.5\baselineskip}

\noindent\textbf{\emph{1) For Grassmannian representations}}

As studied in \cite{edelman1998geometry, hamm2008gda, hamm2008extended, harandi2011ggda, vemulapalli2013kernel}, each Riemannian representation on a Grassmann manifold $\mathcal{G}(d,D)$ refers to a $d$-dimensional linear subspace of $\mathbb{R}^D$. The linear subspace can be represented by its orthonormal basis matrix $\bm{U}$ that is formed by the $d$ leading eigenvectors corresponding to the $d$ largest eigenvalues of the covariance matrix of one Euclidean data set. On a Grassmann manifold, one of the most popular Riemannian metrics is the projection metric \cite{edelman1998geometry}. Formally, for one pair of data $\bm{y}_i,\bm{y}_j$ on the Grassmannian, their distance is measured by the projection metric:
\begin{equation}
d(\bm{y}_i,\bm{y}_j) = 2^{-1/2}\|\bm{U}_i\bm{U}_i^T-\bm{U}_j\bm{U}_j^T\|_\mathcal{F},
\label{K_Eq7}
\end{equation}
where $\|\cdot\|_\mathcal{F}$ denotes the matrix Frobenius norm.


\setlength{\parskip}{0.5\baselineskip}

\noindent\textbf{\emph{2) For affine Grassmannian representations}}

In contrast to the Grassmannian representation, each affine Grassmannian point is an element on an affine Grassmann manifold, which is the space of $d$-dimensional affine subspaces named affine Grassmann manifold $\mathcal{AG}(d,D)$. Therefore, each Riemannian representation on $\mathcal{AG}(d,D)$ is an affine subspace spanned by an orthonormal matrix $\bm{U}$ adding the offset $\bm{\mu}$ (i.e., the mean) from the origin. On the affine Grassmann manifold, \cite{hamm2008extended} defined a similarity function as $tr(\bm{U}_i\bm{U}_i^T\bm{U}_j\bm{U}_i^T)+\bm{\mu}_i^T(\bm{I}-\bm{U}_i\bm{U}_i^T)(\bm{I}-\bm{U}_j\bm{U}_j^T)\bm{\mu}_j$ between pairs of data points. Alternatively, we extend the similarity function to a distance metric between two Riemannian data $\bm{y}_i,\bm{y}_j$ on the affine manifold as:
\begin{equation}
\begin{aligned}
d(\bm{y}_i,\bm{y}_j) &= 2^{-1/2}(\|\bm{U}_i\bm{U}_i^T-\bm{U}_j\bm{U}_j^T\|_\mathcal{F}\\
& +\|(\bm{I}-\bm{U}_i\bm{U}_i^T)\bm{\mu}_i-(\bm{I}-\bm{U}_j\bm{U}_j^T)\bm{\mu}_j\|_\mathcal{F}),
\end{aligned}
\label{K_Eq8}
\end{equation}
where $\bm{I} \in \mathbb{R}^{D \times D}$ is the identity matrix.

\setlength{\parskip}{0.5\baselineskip}

\noindent\textbf{\emph{3) For SPD representations}}

Each SPD representation is an element of the manifold of Symmetric Positive Definite (SPD) matrices $\bm{C}$ of size $D \times D$. As studied in \cite{pennect2006aid, arsigny2007led, wang2012covariance, jayasumana2015kernel}, the set of SPD matrices yields a Riemannian manifold $\mathbb{S}_{+}^{D}$ when endowing a specific Riemannian metric. One of the most commonly used SPD Riemannian metrics is the Log-Euclidean metric \cite{arsigny2007led} due to its effectiveness in encoding the true Riemannian geometry by reducing the manifold to a flat tangent space at the identity matrix. Formally, on the Riemannian SPD manifold, the Log-Euclidean distance metric between two elements $\bm{y}_i,\bm{y}_j$ is given by classical Euclidean computations in the domain of SPD matrix logarithms as:
\begin{equation}
d(\bm{y}_i,\bm{y}_j) = \|\log(\bm{C}_i)-\log(\bm{C}_j)\|_\mathcal{F},
\label{K_Eq9}
\end{equation}
where $\log(\bm{C})=\bm{U}\log(\bm{\Sigma})\bm{U}^T$ with $\bm{C}=\bm{U\Sigma U}^T$ being the eigen-decomposition of the SPD matrix $\bm{C}$.

Similar to our prior work \cite{huang2014lerm}, we denote the proposed CERML working in the three studied settings by CERML-EG, CERML-EA and CERML-ES, respectively. By studying the Riemannian metrics defined in Eq.\ref{K_Eq7}, Eq.\ref{K_Eq8} and Eq.\ref{K_Eq9}, the kernel function corresponding to the specific type of Riemannian manifold can be derived by employing Eq.\ref{K_Eq6}. However, according to Mercer's theorem, only positive definite kernels yield valid RKHS. To achieve this, by employing the approach developed in \cite{jayasumana2015kernel}, we can easily prove the positive definiteness of these Gaussian kernels defined on the resulting Riemannian manifolds. As for the details to prove their positive definiteness, readers are referred to \cite{jayasumana2015kernel}.


\subsection{Objective Function} \label{func}

From Eq.\ref{Eq2}, Eq.\ref{Eq3}, Eq.\ref{Eq4}, we find that the CERML contains two parameter transformation matrices $\bm{W}_x, \bm{W}_y$. In order to learn a discriminant metric between heterogeneous data, we formulate the objective function of this new framework to optimize the two matrices $\bm{W}_x, \bm{W}_y$ in the following:
\begin{equation}
\begin{aligned}
& \mathop{\min}_{\bm{W}_x, \bm{W}_y} J(\bm{W}_x, \bm{W}_y)  \\ & = \mathop{\min}_{\bm{W}_x, \bm{W}_y} \left \{D(\bm{W}_x, \bm{W}_y)
 + \lambda_1 G(\bm{W}_x, \bm{W}_y)+\lambda_2 T(\bm{W}_x, \bm{W}_y)\}, \right.
\label{Eq5}
\end{aligned}
\end{equation}
where $D(\bm{W}_x,\bm{W}_y)$ is the distance constraint defined on the collections of similarity and dissimilarity constraints. $G(\bm{W}_x,\bm{W}_y)$ and $T(\bm{W}_x,\bm{W}_y)$ are, respectively, a geometry constraint and a transformation constraint, both of which are regularizations defined on the target transformation matrices $\bm{W}_x, \bm{W}_y$. $\lambda_1 > 0, \lambda_2 > 0$ are balancing parameters.

\textbf{Distance constraint $D(\bm{W}_x,\bm{W}_y)$}: This constraint is defined so that the distances between the Euclidean data and the Riemannian data -- with the similarity (/dissimilarity) constraints -- are minimized (/maximized). In this paper, we adopt a classical expression of the sum of squared distances to define this constraint as:
\begin{equation}
\begin{aligned}
D(\bm{W}_x,\bm{W}_y) &= \frac{1}{2} \sum_{i=1}^{m} \sum_{j=1}^{n} \bm{A}(i,j)d^2(\bm{x}_i, \bm{y}_j),\\
\bm{A}(i,j) &= \begin{cases}
1, &\text{if $l^x_i = l^y_j$},\\
-1, &\text{if $l^x_i \neq l^y_j$},
\end{cases}
\label{Eq6}
\end{aligned}
\end{equation}
where $\bm{A}(i,j)$ indicates if the heterogeneous data $\bm{x}_i$ and $\bm{y}_j$ are relevant or irrelevant, as inferred from the class label. To balance the effect of similarity and dissimilarity constraints, we normalize the elements of $\bm{A}$ by averaging them over the total number of similar/dissimilar pairs respectively.

\textbf{Geometry constraint $G(\bm{W}_x,\bm{W}_y)$}: This \mbox{constraint} aims to preserve Euclidean geometry and Riemannian geometry for the Euclidean and Riemannian points, respectively. Thus, it can be defined as: $G(\bm{W}_x,\bm{W}_y) = G_x(\bm{W}_x)+G_y(\bm{W}_y)$, which are Euclidean and Riemannian geometry \mbox{preserving} items formulated as:
\begin{equation}
\begin{aligned}
G_x(\bm{W}_x) &= \frac{1}{2} \sum_{i=1}^{m} \sum_{j=1}^{m} \bm{A}_x(i,j)d_x^2(\bm{x}_i, \bm{x}_j),\\
\bm{A}_x(i,j) &=
\begin{cases}
a_{ij}, &\text{if $l^x_i = l^x_j$ and $k_1(i,j)$},\\
-a_{ij}, &\text{if $l^x_i \neq l^x_j$ and $k_2(i,j)$},\\
0, &\text{else},
\end{cases}
\label{Eq7}
\end{aligned}
\end{equation}
\begin{equation}
\begin{aligned}
G_y(\bm{W}_y) &= \frac{1}{2} \sum_{i=1}^{n} \sum_{j=1}^{n} \bm{A}_y(i,j)d_y^2(\bm{y}_i, \bm{y}_j),\\
\bm{A}_y(i,j) &=
\begin{cases}
a_{ij}, &\text{if $l^y_i = l^y_j$ and $k_1(i,j)$},\\
-a_{ij}, &\text{if $l^y_i \neq l^y_j$ and $k_2(i,j)$},\\
0, &\text{else},
\end{cases}
\label{Eq8}
\end{aligned}
\end{equation}
where $a_{ij}=exp(\|\hat{\bm{x}}_i-\hat{\bm{x}}_j\|^2/\sigma^2)$, $\hat{\bm{x}}$ indicates Euclidean data $\bm{x}$ or Riemannian data $\bm{y}$. $k_1(i,j)$ ($k_2(i,j)$) means data $z_i$ is in the $k_1$ ($k_2$) neighborhood of data $z_j$ or data $z_j$ is in the $k_1$ ($k_2$) neighborhood of data $z_i$.

\textbf{Transformation constraint $T(\bm{W}_x, \bm{W}_y)$}: Since Euclidean distance will be used in the target common subspace where all dimensions are treated uniformly, it is reasonable to require the feature vectors satisfy an isotropic distribution. Thus, this constraint can be expressed in terms of unit covariance:
\begin{equation}
T(\bm{W}_x, \bm{W}_y) = \frac{1}{2} (\|\bm{W}^T_x \bm{K}_{x}\|_{\mathcal{F}}^2+\|\bm{W}^T_y \bm{K}_{y}\|_{\mathcal{F}}^2),
\label{Eq9}
\end{equation}
where $\|\cdot\|_{\mathcal{F}}$ is the Frobenius norm.

\subsection{Optimization Algorithm} \label{opt}

To optimize the objective function Eq.\ref{Eq5}, we develop an iterative optimization algorithm, which first applies the Fisher criterion of Fisher Discriminant Analysis (FDA) \cite{fisher1936use} to initialize the two transformation matrices $\bm{W}_x, \bm{W}_y$, and then employs a strategy of alternately updating their values.

Before introducing the optimization algorithm, we first rewrite Eq.\ref{Eq6}, Eq.\ref{Eq7} and \mbox{Eq.\ref{Eq8}} in matrix formulation as:
\begin{equation}
\begin{aligned}
D(\bm{W}_x,\bm{W}_y) &= \frac{1}{2} (\bm{W}_x^T\bm{K}_{x}\bm{B}^{'}_x\bm{K}_{x}^T\bm{W}_x+\bm{W}_y^T\bm{K}_{y}\bm{B}^{'}_y\bm{K}_{y}^T\bm{W}_y\\
&-2\bm{W}_x^T\bm{K}_{x}\bm{A}\bm{K}^T_{y}\bm{W}_y),
\label{Eq10}
\end{aligned}
\end{equation}
\begin{equation}
\begin{aligned}
G_x(\bm{W}_x) &= \bm{W}_x^T\bm{K}_{x}\bm{B}_x\bm{K}_{x}^T\bm{W}_x-\bm{W}_x^T\bm{K}_{x}\bm{A}_x\bm{K}_{x}^T\bm{W}_x\\
& = \bm{W}_x^T\bm{K}_{x}\bm{L}_x\bm{K}_{x}^T\bm{W}_x,
\label{Eq11}
\end{aligned}
\end{equation}
\begin{equation}
\begin{aligned}
G_y(\bm{W}_y) &= \bm{W}_y^T\bm{K}_{y}\bm{B}_y\bm{K}_{y}^T\bm{W}_y-\bm{W}_y^T\bm{K}_{y}\bm{A}_y\bm{K}_{y}^T\bm{W}_y\\
& = \bm{W}_y^T\bm{K}_{y}\bm{L}_y\bm{K}_{y}^T\bm{W}_y,
\label{Eq12}
\end{aligned}
\end{equation}
where $\bm{B}^{'}_x$, $\bm{B}^{'}_y$, $\bm{B}_x$ and $\bm{B}_y$ are diagonal matrices with $\bm{B}^{'}_x(i,i)=\sum_{j=1}^n \bm{A}(i,j)$, $\bm{B}^{'}_y(j,j)=\sum_{i=1}^m \bm{A}(i,j)$, $\bm{B}_x(i,i)=\sum_{j=1}^m \bm{A}_x(i,j)$, $\bm{B}_y(i,i)=\sum_{j=1}^n \bm{A}_y(i,j)$.

\textbf{Initialization}. We define the within-class template $\bm{A}^w$ and between-class template $\bm{A}^b$ for $\bm{A}$ in Eq.\ref{Eq6} as:
\begin{equation}
\bm{A}^w(i,j) = \begin{cases}
1, &\text{if $l^x_i = l^y_j$},\\
0, &\text{if $l^x_i \neq l^y_j$},
\end{cases}
\bm{A}^b(i,j) = \begin{cases}
0, &\text{if $l^x_i = l^y_j$},\\
1, &\text{if $l^x_i \neq l^y_j$}.
\end{cases}
\label{Eq13}
\end{equation}

By substituting Eq.\ref{Eq13} into Eq.\ref{Eq10}, the within-class template $D^w(\bm{W}_x,\bm{W}_y)$ and between-class template $D^b(\bm{W}_x,\bm{W}_y)$ for $D(\bm{W}_x,\bm{W}_y)$ in Eq.\ref{Eq6} can be computed as:
\begin{equation}
\begin{aligned}
D^w(\bm{W}_x,\bm{W}_y) & = \frac{1}{2} (\bm{W}_x^T\bm{K}_{x}\bm{B}^{'w}_x\bm{K}_{x}^T\bm{W}_x
+\bm{W}_y^T\bm{K}_{y}\bm{B}^{'w}_y\bm{K}_{y}^T\bm{W}_y \\&-2\bm{W}_x^T\bm{K}_{x}\bm{Z}^w\bm{K}_{y}\bm{W}_y^T),
\label{Eq19}
\end{aligned}
\end{equation}
\begin{equation}
\begin{aligned}
D^b(\bm{W}_x,\bm{W}_y) & = \frac{1}{2} (\bm{W}_x^T\bm{K}_{x}\bm{B}^{'b}_x\bm{K}_{x}^T\bm{W}_x
+\bm{W}_y^T\bm{K}_{y}\bm{B}^{'b}_y\bm{K}_{y}^T\bm{W}_y \\&-2\bm{W}_x^T\bm{K}_{x}\bm{Z}^b\bm{K}_{y}\bm{W}_y^T).
\label{Eq20}
\end{aligned}
\end{equation}

Likewise, we achieve the within-class and between-class templates for $G_x$ and $G_{y}$ in Eq.\ref{Eq7} and Eq.\ref{Eq8} respectively denoted by $G^w_x$, $G^b_x$, $G^w_{y}$, $G^b_{y}$. For the sake of clarity, more details are given in the supplementary material.

Then we can initialize $\bm{W}_x$ and $\bm{W}_{y}$ by maximizing the sum of between-class templates while minimizing the sum of within-class templates with the Fisher criterion of the traditional Fisher Discriminant Analysis (FDA) \cite{fisher1936use}:
\begin{equation}
\begin{aligned}
 \mathop{\max}_{\bm{W}_x,\bm{W}_y} & \{\bm{D}^b(\bm{W}_x,\bm{W}_y) + \lambda_1 G^b(\bm{W}_x,\bm{W}_y)\},\\
 s.t. \quad & \bm{D}^w(\bm{W}_x,\bm{W}_y) + \lambda_1 G^w(\bm{W}_x,\bm{W}_y) = 1,
\label{Eq25}
\end{aligned}
\end{equation}
where $G^b(\bm{W}_x, \bm{W}_y)=G^b_x(\bm{W}_x, \bm{W}_y)+G^b_y(\bm{W}_x, \bm{W}_y)$, $G^w(\bm{W}_x, \bm{W}_y)=G^w_x(\bm{W}_x, \bm{W}_y)+G^w_y(\bm{W}_x, \bm{W}_y)$. By transforming Eq.\ref{Eq25} into matrix formulation, the function for initialization can be further simplified as:
\begin{equation}
\begin{aligned}
  &\max \bm{W}^T\bm{M}^b\bm{W}, \quad s.t.\bm{W}^T\bm{M}^w\bm{W}= 1, \\
  &\Rightarrow \bm{M}^b\bm{W} = \lambda \bm{M}^w\bm{W},
\label{Eq27}
\end{aligned}
\end{equation}
where $\bm{M}^b=\begin{bmatrix}\bm{K}_{x}\bm{R}_x^{b}\bm{K}_{x}^T & -\bm{K}_{x}\bm{Z}^b\bm{K}_{y}^T\\-\bm{K}_{y}({\bm{Z}^b})^T \bm{K}_{x}^T & \bm{yR}_y^{b}\bm{K}_{y}^T\\\end{bmatrix},\bm{M}^w=\begin{bmatrix}\bm{K}_{x}\bm{R}_x^{w}\bm{K}_{x}^T & -\bm{K}_{x}\bm{Z}^w\bm{K}_{y}^T\\-\bm{K}_{y}({\bm{Z}^w})^T \bm{K}_{x}^T & \bm{yR}_y^{w}\bm{K}_{y}^T\\\end{bmatrix}$, $\bm{W}^T=[\bm{W}_x^T, \bm{W}_y^T]$. Obviously, Eq.\ref{Eq27} is a standard generalized eigenvalue problem that can be solved using any eigensolver.

\textbf{Alternately updating}. We substitute Eq.\ref{Eq10}, Eq.\ref{Eq11}, Eq.\ref{Eq12} into the objective function $J(\bm{W}_x, \bm{W}_y)$ in Eq.\ref{Eq5} to derive its matrix form. By differentiating $J(\bm{W}_x, \bm{W}_y)$ w.r.t. $\bm{W}_x$ and setting it to zero, we have the following equation:
\begin{equation}
\begin{aligned}
\frac{\partial Q(\bm{W}_x,\bm{W}_y)}{\partial \bm{W}_x} &= \bm{K}_{x}\bm{B}^{'}_x\bm{K}_{x}^T\bm{W}_x-\bm{K}_{x}\bm{Z}\bm{K}_{y}^T\bm{W}_y \\
& +2\lambda_1 \bm{K}_{x}\bm{L}_x\bm{K}_{x}^T\bm{W}_x+2\lambda_2\bm{K}_{x}\bm{K}_{x}^T\bm{W}_x = 0.
\label{Eq28}
\end{aligned}
\end{equation}
Then by fixing $\bm{W}_y$, the solution of $\bm{W}_x$ can be achieved as:
\begin{equation}
\bm{W}_x = (\bm{K}_{x}(\bm{B}^{'}_x+2 \lambda_1 \bm{L}_x+2\lambda_2\bm{I})\bm{K}_{x}^T)^{-1}\bm{K}_{x}\bm{A}\bm{K}_{y}^T\bm{W}_y.
\label{Eq29}
\end{equation}
Likewise, the solution of $\bm{W}_y$ when $\bm{W}_x$ is fixed, can be obtained as
\begin{equation}
\bm{W}_y = (\bm{K}_{y}(\bm{B}^{'}_y+2 \lambda_1 \bm{L}_y+2\lambda_2\bm{I})\bm{K}_{y}^T)^{-1}\bm{K}_{y}\bm{A}\bm{K}_{x}^T\bm{W}_x.
\label{Eq30}
\end{equation}

We alternate the above updates of $\bm{W}_x$ and $\bm{W}_y$ for several iterations to search an optimal solution. While it is hard to provide a theoretical proof of uniqueness or convergence of the proposed iterative optimization, we empirically found our objective function Eq.\ref{Eq5} can converge to a desirable solution after only a few tens of iterations. The convergence characteristics are studied in more detail in the experiments.

\section{Application to Video-based Face Recognition} \label{appli}

In this section we present the application of the proposed Cross Euclidean-to-Riemannian Metric Learning (CERML) to the three typical tasks of video-based face recognition, i.e., V2S, S2V and V2V settings.

\subsection{V2S/S2V Face Recognition} \label{v2s/s2v}

As done in several state-of-the-art techniques \cite{hamm2008gda, turaga2011statistical, wang2012covariance, lu2013image,  harandi2014manifold, huang2015face, wang2015discriminant}, we represent a set of facial frames within one video with their appearance mean and variation model aforementioned (e.g., linear subspace) simultaneously. Therefore, the task of V2S/S2V face recognition can be formulated as the problem of matching Euclidean representations (i.e., feature vectors) of face images with the Euclidean data (i.e., feature mean) and Riemannian representation (i.e., feature variation) of faces from videos. Formally, the Euclidean data of a face image is written as $\bm{X}=\{\bm{x}_1,\bm{x}_2,\ldots,\bm{x}_m\}, \bm{x}_i \in \mathbb{R}^{D_1}$ with labels $\{l^x_1,l^x_2,\ldots,l^x_m\}$. The Euclidean data of videos are represented by $\bm{Y}=\{\bm{y}_1,\bm{y}_2,\ldots,\bm{y}_n\}, \bm{y}_j \in \mathbb{R}^{D_2}$, with labels $\{l^y_1,l^y_2,\ldots,l^y_n\}$, while their Riemannian data are $\bm{Z}=\{\bm{z}_1,\bm{z}_2,\ldots,\bm{z}_n\}, \bm{z}_j \in \mathcal{M}$ sharing the labels with their Euclidean data. In the following, we describe the components of the proposed CERML framework for this task.

\textbf{Distance metric}. The distance metric Eq.\ref{Eq2} in Sec.\ref{form} is instantiated for V2S/S2V face recognition as:
\begin{equation}
\begin{aligned}
& d(\bm{x}_i, \bm{y}_j) +d\bm{x}_i, \bm{z}_j) \\
 & = \sqrt{(\bm{W}^T_x \bm{K}_{x_{.i}}-\bm{W}^T_y \bm{K}_{y_{.j}})^T(\bm{W}^T_x \bm{K}_{x_{.i}}-\bm{W}^T_y \bm{K}_{y_{.j}})} \\
& +\sqrt{(\bm{W}^T_x \bm{K}_{x_{.i}}-\bm{W}^T_z \bm{K}_{z_{.j}})^T(\bm{W}^T_x \bm{K}_{x_{.i}}-\bm{W}^T_z \bm{K}_{z_{.j}})}.
\end{aligned}
\label{Eq31}
\end{equation}

\textbf{Objective function}. The objective function Eq.\ref{Eq5} in Sec.\ref{opt} takes the form:
\begin{equation}
\begin{aligned}
& \mathop{\min}_{\bm{W}_x, \bm{W}_y, \bm{W}_z} J(\bm{W}_x, \bm{W}_y,\bm{W}_z) \\
& = \mathop{\min}_{\bm{W}_x, \bm{W}_y, \bm{W}_z} \{D(\bm{W}_x, \bm{W}_y, \bm{W}_z) + \lambda_1 G(\bm{W}_x, \bm{W}_y, \bm{W}_z) \\ & +\lambda_2 T(\bm{W}_x, \bm{W}_y,\bm{W}_z)\},
\end{aligned}
\label{Eq32}
\end{equation}
where the distance constraint $D(\bm{W}_x, \bm{W}_y, \bm{W}_z)=D(\bm{W}_x, \bm{W}_y)+D(\bm{W}_x, \bm{W}_z)$, the geometry constraint $G(\bm{W}_x, \bm{W}_y, \bm{W}_z) =G_x(\bm{W}_x)+G_y(\bm{W}_y)+G_z(\bm{W}_z)$, the transformation constraint $T(\bm{W}_x, \bm{W}_y,\bm{W}_z)=\frac{1}{2} (\|\bm{W}^T_x \bm{K}_{\bm{X}}\|_{\mathcal{F}}^2+\|\bm{W}^T_y \bm{K}_{\bm{Y}}\|_{\mathcal{F}}^2+\|\bm{W}^T_z \bm{K}_{\bm{Z}}\|_{\mathcal{F}}^2)$.

\textbf{Initialization}. In the initialization function Eq.\ref{Eq27}, the optimization algorithm in Sec.\ref{opt} is instantiated with the matrix $\bm{M}^b, \bm{M}^w$ as:
\begin{equation}
\bm{M}^b = \begin{bmatrix}\bm{K}_{x}\bm{R}_x^{b}\bm{K}_{x}^T & -\bm{K}_{x}\bm{A}_{xy}^b\bm{K}_{y}^T & -\bm{K}_{x}\bm{A}_{xz}^b\bm{K}_{z}^T\\
 -\bm{K}_{y}({\bm{A}_{xy}^b})^T \bm{K}_{x}^T & \bm{K}_y\bm{R}_y^{b}\bm{K}_{y}^T & 0 \\
 -\bm{K}_{z}({\bm{A}_{xz}^b})^T \bm{K}_{x}^T  & 0 & \bm{K}_z\bm{R}_z^{b}\bm{K}_{z}^T\\
\end{bmatrix},
\label{Eq34}
\end{equation}
\begin{equation}
 \bm{M}^w=\begin{bmatrix}\bm{K}_{x}\bm{R}_x^{w}\bm{K}_{x}^T & -\bm{K}_{x}\bm{A}_{xy}^w\bm{K}_{y}^T & -\bm{K}_{x}\bm{A}_{xz}^w\bm{K}_{z}^T\\
 -\bm{K}_{y}({\bm{A}_{xy}^w})^T \bm{K}_{x}^T & \bm{K}_y\bm{R}_y^{w}\bm{K}_{y}^T & 0\\
 -\bm{K}_{z}({\bm{A}_{xz}^w})^T \bm{K}_{x}^T & 0 & \bm{K}_z\bm{R}_z^{w}\bm{K}_{z}^T \\
 \end{bmatrix}.
\label{Eq35}
\end{equation}

\textbf{Alternately updating}. The analytical solutions Eq.\ref{Eq29} and Eq.\ref{Eq30} in the Sec.\ref{opt} can be rewritten as:
\begin{equation}
\begin{aligned}
\bm{W}_x & = (\bm{K}_{x}(2\bm{B}^{'}_x+2 \lambda_1 \bm{L}_x+2\lambda_2\bm{I}) \bm{K}_{x}^T)^{-1} \\ & (\bm{K}_{x}\bm{A}_{xy}\bm{K}_{y}^T\bm{W}_y
+\bm{K}_{x}\bm{A}_{xz}\bm{K}_{z}^T\bm{W}_z), \\
\bm{W}_y & = (\bm{K}_{y}(\bm{B}^{'}_y+2 \lambda_1 \bm{L}_y+2\lambda_2\bm{I})\bm{K}_{y}^T)^{-1}\bm{K}_{y}\bm{A}_{xy}\bm{K}_{x}^T\bm{W}_x, \\
\bm{W}_z & = (\bm{K}_{z}(\bm{B}^{'}_z+2 \lambda_1 \bm{L}_z+2\lambda_2\bm{I})\bm{K}_z^T)^{-1}\bm{K}_{z}\bm{A}_{xz}\bm{K}_{x}^T\bm{W}_x.
\label{Eq36}
\end{aligned}
\end{equation}

\subsection{V2V Face Recognition} \label{v2v}

Similar to the case of V2S/S2V face recognition, each facial video sequence is commonly represented by the appearance mean of its frames and their pattern variation. Therefore, the task of V2S/S2V face recognition can be expressed as the problem of fusing the Euclidean data (i.e., feature mean) and the Riemannian representation (i.e., feature variation such as linear subspace) of video sequences of faces. Formally, the Euclidean data of videos are represented by $\bm{Y}=\{\bm{y}_1,\bm{y}_2,\ldots,\bm{y}_n\}, \bm{y}_j \in \mathbb{R}^{D_2}$, with labels $\{l^y_1,l^y_2,\ldots,l^y_n\}$, while the Riemannian representations of such videos are $\bm{Z}=\{\bm{z}_1,\bm{z}_2,\ldots,\bm{z}_n\}, \bm{z}_j \in \mathcal{M}$ sharing the labels with the Euclidean data. To adapt the proposed CERML framework to this task, we now define its components.

\textbf{Distance metric}. The distance metrics Eq.\ref{Eq3} and Eq.\ref{Eq4} in Sec.\ref{form} are implemented for V2V face recognition as:
\begin{equation}
\begin{aligned}
& d(\bm{y}_i, \bm{y}_j) +d(\bm{z}_i, \bm{z}_j) \\
& =\sqrt{(\bm{W}^T_y \bm{K}_{y_{.i}}-\bm{W}^T_y \bm{K}_{y_{.j}})^T(\bm{W}^T_y \bm{K}_{y_{.i}}-\bm{W}^T_y \bm{K}_{y_{.j}})}\\
& +\sqrt{(\bm{W}^T_z \bm{K}_{z_{.i}}-\bm{W}^T_z \bm{K}_{z_{.j}})^T(\bm{W}^T_z \bm{K}_{z_{.i}}-\bm{W}^T_z \bm{K}_{z_{.j}})}.
\end{aligned}
\label{Eq37}
\end{equation}

\textbf{Objective function}. The objective function Eq.\ref{Eq5} in Sec.\ref{opt} is instantiated as:
\begin{equation}
\begin{aligned}
&\mathop{\min}_{\bm{W}_y, \bm{W}_z} J(\bm{W}_y,\bm{W}_z)\\
& =\mathop{\min}_{\bm{W}_y,\bm{W}_z} \{D(\bm{W}_y, \bm{W}_z) + \lambda_1 G(\bm{W}_y, \bm{W}_z) +\lambda_2 T(\bm{W}_y,\bm{W}_z)\},
\end{aligned}
\label{Eq38}
\end{equation}
where the distance constraint $D(\bm{W}_y, \bm{W}_z)=D(\bm{W}_y)+D(\bm{W}_z)$, the geometry constraint $G(\bm{W}_y, \bm{W}_z) =G_y(\bm{W}_y)+G_z(\bm{W}_z)$, the transformation constraint $T(\bm{W}_y,\bm{W}_z)=\frac{1}{2} (\|\bm{W}^T_y \bm{K}_{\bm{Y}}\|_{\mathcal{F}}^2+\|\bm{W}^T_z \bm{K}_{\bm{Z}}\|_{\mathcal{F}}^2)$.

\textbf{Initialization}. The initialized objective function Eq.\ref{Eq27} in the optimization algorithm in Sec.\ref{opt} is instantiated by defining the matrix $\bm{M}^b, \bm{M}^w$ as:
\begin{equation}
\bm{M}^b = \begin{bmatrix}\bm{K}_{y}\bm{R}_y^{b}\bm{K}_{y}^T & -\bm{K}_{y}\bm{A}^b\bm{K}_{z}^T\\-\bm{K}_{z}({\bm{A}^b})^T \bm{K}_{y}^T & \bm{K}_z\bm{R}_z^{b}\bm{K}_{z}^T\\\end{bmatrix},
\label{Eq40}
\end{equation}

\begin{equation}
 \bm{M}^w=\begin{bmatrix}\bm{K}_{y}\bm{R}_y^{w}\bm{K}_{y}^T & -\bm{K}_{y}\bm{A}^w\bm{K}_{z}^T\\-\bm{K}_{z}({\bm{A}^w})^T \bm{K}_{y}^T & \bm{K}_z\bm{R}_z^{w}\bm{K}_{z}^T\\\end{bmatrix}.
\label{Eq41}
\end{equation}

\textbf{Alternately updating}. The analytical solutions Eq.\ref{Eq29} and Eq.\ref{Eq30} in Sec.\ref{opt} can be derived as:
\begin{equation}
\begin{aligned}
\bm{W}_y & = (\bm{K}_{y}(\bm{B}^{'}_y+2 \lambda_1 \bm{L}_y+2\lambda_2\bm{I})\bm{K}_{y}^T)^{-1}\bm{K}_{y}\bm{A}\bm{K}_{z}^T\bm{W}_z, \\
\bm{W}_z & = (\bm{K}_z(\bm{B}^{'}_z+2 \lambda_1 \bm{L}_z+2\lambda_2\bm{I})\bm{K}_z^T)^{-1}\bm{K}_{z}\bm{A}\bm{K}_{y}^T\bm{W}_y.
\label{Eq42}
\end{aligned}
\end{equation}

\begin{figure}[t]
\begin{center}
\includegraphics[width=0.95\linewidth]{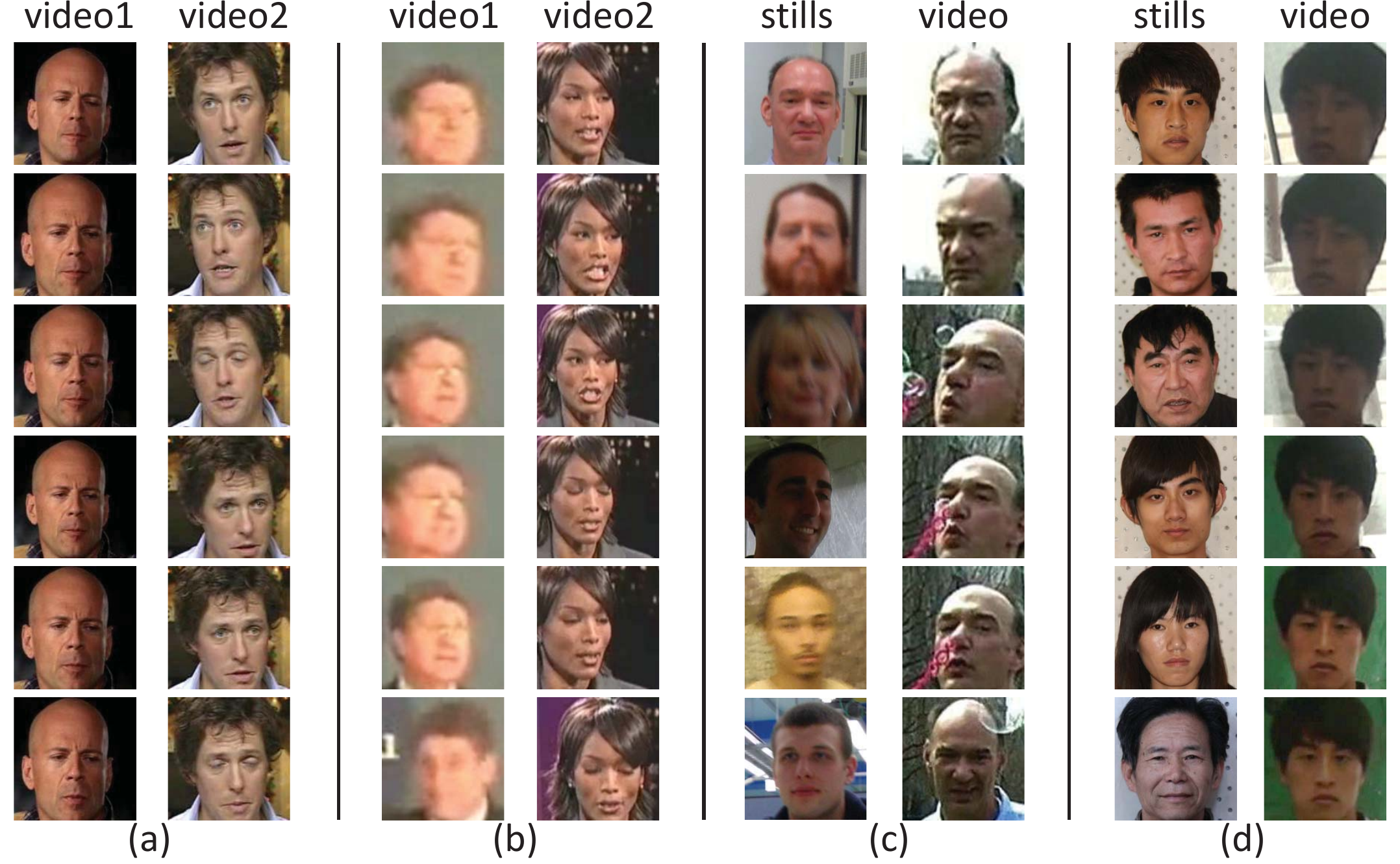}

\end{center}
\caption{Example still images and video frames from two internet video face databases YTC (a), YTF (b) and two surveillance-like video face databases PaSC (c), COX (d).}
\label{Fig4}
\end{figure}

\section{Experimental Evaluation}

To study the effectiveness of the proposed CERML, we conduct evaluations for the three typical tasks of video-based face recognition being -- Video-to-Still (V2S), Still-to-Video (S2V) and Video-to-Video (V2V) face recognition -- on four challenging video face databases. These databases are YouTube Celebrities (YTC) \cite{kim2008ytc}, YouTube Face DB (YTF) \cite{wolf2011youtube}, Point-and-Shoot Challenge (PaSC) \cite{beveridge2013challenge} and COX face database \cite{huang2015cox}.

The YTC \cite{kim2008ytc} and the YTF \cite{wolf2011youtube} are both popular internet video face databases. Both of them are collected from YouTube. The YTC database consists of 1,910 video clips of 47 different celebrities. The video frames often exhibit a large variation of pose and illumination, as well as degradations such as noise or compression effects (see Fig.\ref{Fig4} (a)). The YTF, another internet video face database, contains 3,425 videos of 1,595 different persons. As shown in Fig.\ref{Fig4} (b), there are again large variations in pose, illumination, and expression. Therefore, the YTF databases are very challenging for verifying faces from these internet videos.

The PaSC \cite{beveridge2013challenge} and COX \cite{huang2015cox} are designed to simulate more of a video surveillance system. The PaSC database \cite{beveridge2013challenge} was collected for the problem of face recognition from stills and videos captured by point-and-shoot cameras in the context of social networks. This database includes 9,376 still images of 293 people balanced in terms of distance to the camera, alternative sensors, frontal versus not-frontal views, and varying location (see Fig.\ref{Fig4} (c)). There are also 2,802 videos for 265 people, a subset of the 293 in the still image portion of the PaSC. The COX captures 3,000 videos and 1,000 still images of 1,000 different persons with 3 different camcorders located in different positions. As shown in Fig.\ref{Fig4} (d), most of the video frames are of low resolution and low quality, with blur, and captured under poor lighting.

\subsection{V2S/S2V Face Recognition}

Video-to-Still (V2S) and Still-to-Video (S2V) face recognition matches still face images against video sequences of faces. To evaluate our proposed method in these two video-based face recognition scenarios, we employ two standard video face databases being PaSC \cite{beveridge2013challenge} and COX \cite{huang2015cox}.

\subsubsection{Dataset Setting}

On the PaSC, the facial region is cropped from the given video frame based on the eye coordinates, as in \cite{beveridge2013challenge}. The cropped facial region is aligned and scaled to a size of $224 \times 224$ pixels. Following the work of others \cite{huang2015face, beveridge2015report, huang2015projection}, we use the approach of \cite{parkhi2015deep} to extract state-of-the-art deep face feature from the normalized face images. In the evaluation of V2S face verification on this database, the target signature set contains still images and the query signature set contains video sequences. The target set includes 4,688 images of 293 people while the query set contains 2,801 videos of 265 people. Since the videos are grouped into the sets of control and handheld videos respectively, the V2S evaluation on the PaSC consists of two tests: Video (control)-Still and Video (handheld)-Still.

For the COX database, we use the face detection and the positions of the eyes provided by the work in \cite{huang2015cox}. In the scenario of V2S/S2V face recognition, we implement two types of feature extractions. In the first setting, we rotate and crop each face image to a normalized image of size $48 \times 60$. On the face images, we extract grayscale features. In the other setting, as done on the PaSC, we also extract state-of-the-art deep face features by the approach of \cite{parkhi2015deep}. In the V2S evaluation, the target set contains still images of the persons with known identities, while the query samples are video clips of faces to be recognized. In contrast, the target set of the S2V scenario conversely contains videos of faces while the queries are still face images. There are 3 testing sets of videos, each of which contains 700 videos from 700 subjects. Therefore, in total 6 tests (i.e., V1-S, V2-S, V3-S, S-V1, S-V2, S-V3, where Vi is the i-th video testing set) are conducted.

\begin{table*}[t]
\linespread{1.2}
\caption{V2S/S2V face recognition results (\%) on PaSC and COX using gray/deep features. Here EG, EA and ES represent Euclidean-to-Grassmannian, Euclidean-to-AffineGrassmannian and Euclidean-to-SPD matching cases, respectively. Con and han mean the control and handheld settings.} 
\footnotesize
\begin{center}
\begin{tabular}{|m{2.5cm}<{\centering}|m{1.45cm}<{\centering}|m{1.45cm}<{\centering}||m{1.4cm}<{\centering}|m{1.4cm}<{\centering}
|m{1.4cm}<{\centering}|m{1.4cm}<{\centering}|m{1.4cm}<{\centering}|m{1.4cm}<{\centering}|}
\hline
\multirow{2}{*}{Methods} & \multicolumn{2}{c||}{PaSC (\emph{deep})} & \multicolumn{6}{c|}{COX (\emph{gray/deep})}\\
\cline{2-9} & V(con)-S & V(han)-S & V1-S & V2-S & V3-S & S-V1 & S-V2 & S-V3 \\
\hline\hline
VGGDeepFace \cite{parkhi2015deep} & 68.62 & 65.29 & --/79.10 & --/77.53 & --/79.03 & --/59.31 & --/65.21 & --/74.29 \\
\hline
NCA \cite{goldberger2004neighbourhood} & 68.91 & 67.48  & 39.14/79.83 & 31.57/74.21 & 57.57/80.69 & 37.71/72.01 & 32.14/69.39  & 58.86/79.51 \\
ITML \cite{davis2007information} & 67.25 & 65.33 & 19.83/82.43 & 18.20/73.49 & 36.63/86.53 & 26.66/73.03 & 25.21/61.57  & 47.57/82.29\\
LFDA \cite{sugiyama2007dimensionality} & \textbf{72.44} & \textbf{69.86} & 21.41/66.58 & 22.17/55.29 & 43.99/72.86 & 40.54/78.25 & 33.90/68.14  & 61.40/84.15\\
LMNN \cite{weinberger2009distance} & 71.12 & 67.12 & 34.44/83.25 & 30.03/72.19 & 58.06/83.25 & 37.84/76.12 & 35.77/70.26  & 63.33/80.92\\

PSDML \cite{zhu2013p2s} & 66.15 & 63.61  & 12.14/65.09 & 9.43/58.16 & 25.43/80.05 & 7.04/54.15 & 4.14/49.16  & 29.86/78.57\\
\hline
KPLS\cite{sharma2011pls}-EG  & 42.29 & 41.21 & 21.83/46.19 & 18.50/44.90 & 30.89/44.67 & 15.01/49.77 & 12.41/50.31 & 25.63/55.07\\
KPLS\cite{sharma2011pls}-EA  & 44.42 & 42.30 & 21.54/47.31 & 19.19/46.94 & 29.41/37.96 & 15.73/45.97 & 12.51/47.10  & 24.54/41.41\\
KPLS\cite{sharma2011pls}-ES  & 44.27 & 42.08 & 20.21/46.26 & 16.21/45.16 & 27.23/42.16 & 14.83/46.21 & 11.61/46.21  & 23.99/43.63\\
KCCA\cite{hardoon2004canonical}-EG  & 58.08 & 55.71  &32.51/73.53 & 28.87/70.94 & 48.43/78.79  & 30.16/73.03 & 27.34/70.60  & 44.91/78.59\\
KCCA\cite{hardoon2004canonical}-EA  & 61.87 & 60.58 & 30.33/75.03 & 28.39/72.34  & 47.74/75.94 & 28.49/74.63 & 26.49/72.26  & 45.21/74.79\\
KCCA\cite{hardoon2004canonical}-ES  & 61.36 & 60.11 & 38.60/80.87 & 33.20/76.63 & 53.26/81.94 & 36.39/80.00 & 30.87/76.76  & 50.96/81.40\\
KGMA\cite{sharma2012gma}-EG  & 60.05 & 58.18 & 32.41/75.24 & 28.96/72.91  & 48.37/79.93 & 30.06/75.19 & 27.57/72.57  & 44.99/80.06\\
KGMA\cite{sharma2012gma}-EA  & 64.87 & 63.58  & 30.60/79.33 & 28.34/76.19  & 47.74/79.94 & 28.54/78.63 & 26.20/76.06  & 45.27/78.54\\
KGMA\cite{sharma2012gma}-ES & 63.76 & 62.32 & 41.89/80.91 & 38.29/76.53 & 52.87/81.90 & 38.03/80.00 & 33.29/76.69  & 50.06/81.41\\
\hline
CERML-EG & 63.34  & 60.92 & 32.63/85.71 & 33.89/82.51 &  49.33/\textbf{87.23} & 43.29/88.80 & 41.19/85.69 & 58.71/\textbf{90.99}\\
CERML-EA &  67.95  & 66.16 & 38.77/\textbf{86.40} & 37.57/\textbf{83.13} &  53.93/86.76 & 43.93/\textbf{88.97} & 41.56/\textbf{85.84} & 57.34/90.26 \\
CERML-ES & 70.64  & 68.91 & 51.41/86.21 &  49.81/82.66 & 64.01/86.64 & 52.39/88.93 & 49.39/85.37  & 65.19/89.64 \\
\hline
\end{tabular}
\end{center}
\label{tab_pasc_cox_v2s_s2v}
\end{table*}

\subsubsection{Method Setting}

In the evaluation of V2S/S2V face recognition, besides to the deep learning method VGGDeepFace \cite{parkhi2015deep} that reported the state of the art on YTF, we also compare our proposed approach with two categories of state-of-the-art metric learning methods as listed in the following. The homogeneous (Euclidean) metric learning approaches learn the single-view metric between Euclidean features of video frames/still images, and adopt the maximal pooling on the matching between images and frames within one video. The heterogeneous metric learning methods learn the cross-view metric among the Riemannian representations of videos, their Euclidean features and the Euclidean features of still images. As KPLS, KCCA and KGMA are all both designed to analyze heterogeneous data, we feed them with our proposed Euclidean-to-Grassmannian (EG), Euclidean-to-AffineGrassmannian (EA) and Euclidean-to-SPD (ES) heterogeneous data, which are also the inputs of our CERML. 

\begin{enumerate}
%

 \item Homogeneous (Euclidean) metric learning methods:

   Neighborhood Components Analysis (NCA) \cite{goldberger2004neighbourhood},  Information-Theoretic Metric Learning
   (ITML) \cite{davis2007information}, Local Fisher Discriminant Analysis (LFDA) \cite{sugiyama2007dimensionality}, Large Margin Nearest Neighbor (LMNN) \cite{weinberger2009distance} and Point-to-Set Distance Metric Learning \mbox{(PSDML) \cite{zhu2013p2s}};

  \item Heterogeneous metric learning methods:

    Kernel Partial Least Squares (KPLS) \cite{sharma2011pls}, Kernel Canonical Correlation Analysis (KCCA) \cite{hardoon2004canonical} and Kernel Generalized Multiview Linear Discriminant Analysis (KGMA) \cite{sharma2012gma}.

\end{enumerate}

For fair comparison, we tune the key parameters of the comparative methods according to the suggestions from the original works. For NCA, the maximum number of line searches is set to 5. For ITML, the lower/upper bound distance is set to the mean distance minus/plus standard variance, and its other parameters are set to the default values designed in its released code. In LFDA, the neighborhood number $k$ is set to $7$. For LMNN, the number of neighborhood is set to $5$, the maximum iteration number is set to $500$, the portion of training samples in validation is $30\%$. For \mbox{PSDML}, we set the regularization parameter as $\kappa=0.8$, the number of negative pairs per sample $k=6$. For KPLS and KGMA, the numbers of the factors are set with the class number of training data. For the proposed method CERML, we set the parameters $\lambda_1=0.01, \lambda_2=0.1$, the neighborhood number $k_1 = 1, k_2 =20$, the kernel widths $\sigma$s are all specified from the mean distances respectively on the training multi-view data, and the number of iterations is set to $20$. As studied in our previous work \cite{huang2014lerm}, we find that adding the cross-view kernels (see the supplementary material for their definitions) into the original single-view kernels does improve the V2S/S2V face recognition. Thus, we use the proposed cross-view kernels in \cite{huang2014lerm} to concatenate the EG/EA/ES kernels for CERML. For the comparisons between single-view and cross-view kernel cases, readers are referred to the supplementary material.


\begin{figure}[t]
\begin{center}
\includegraphics[width=0.9\linewidth]{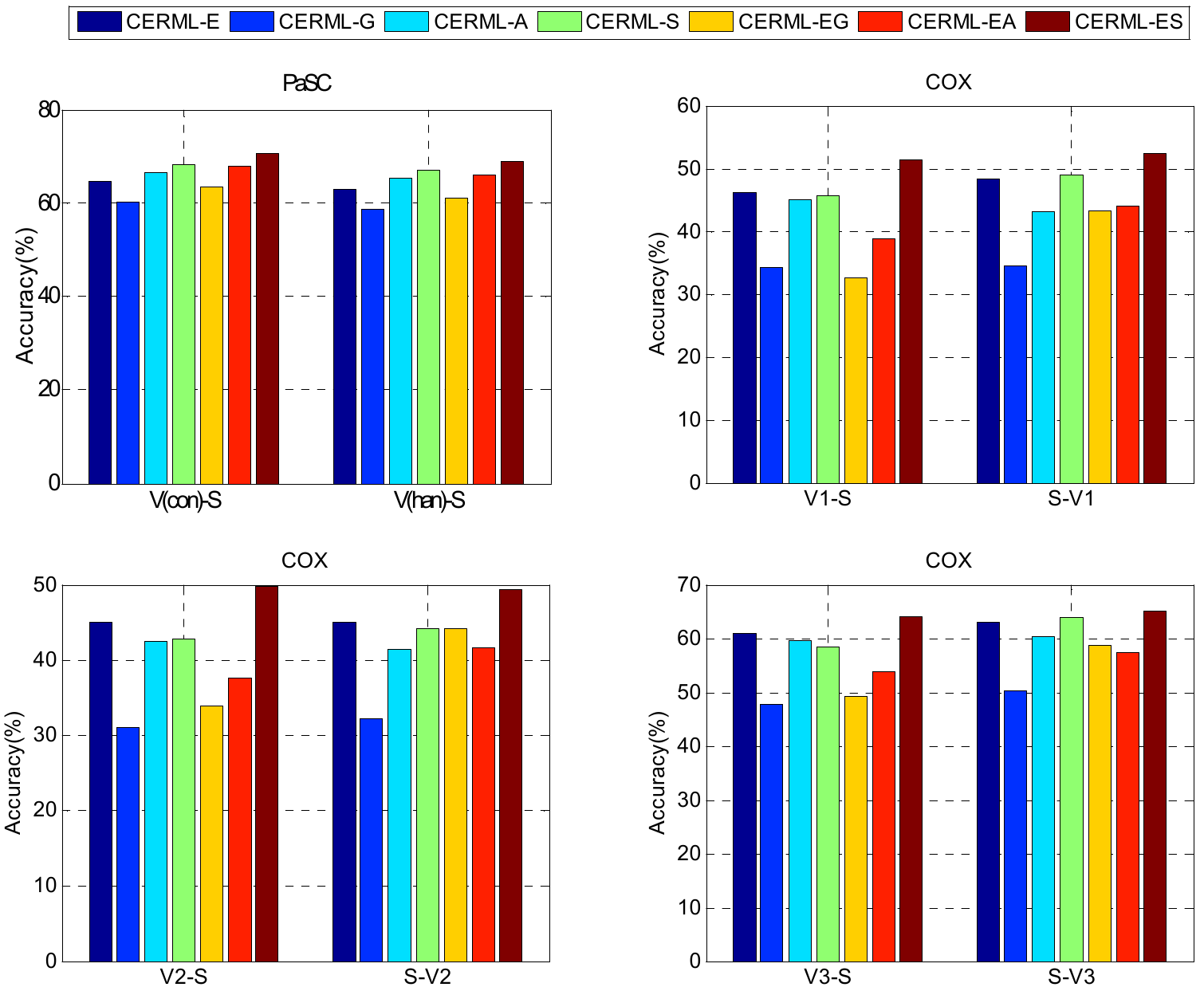}

\end{center}
\caption{V2S/S2V face recognition results (\%) of the proposed CEMRL dealing with different representations of videos for PaSC (deep feature) and COX (gray feature). Here, CERML-E, CERML-G, CERML-A, CERML-S, CERML-EG, CERML-EA, CERML-ES respectively indicate videos are represented by mean, subspace, affine subspace, SPD matrix, mean+subspace, mean+affine subspace, mean+SPD matrix. \emph{Note that CERML-G/A/S is the proposed method in our prior work \cite{huang2014lerm}}.}
\label{Fig5}
\end{figure}

\begin{figure}[t]
\begin{center}
   \includegraphics[width=0.8\linewidth]{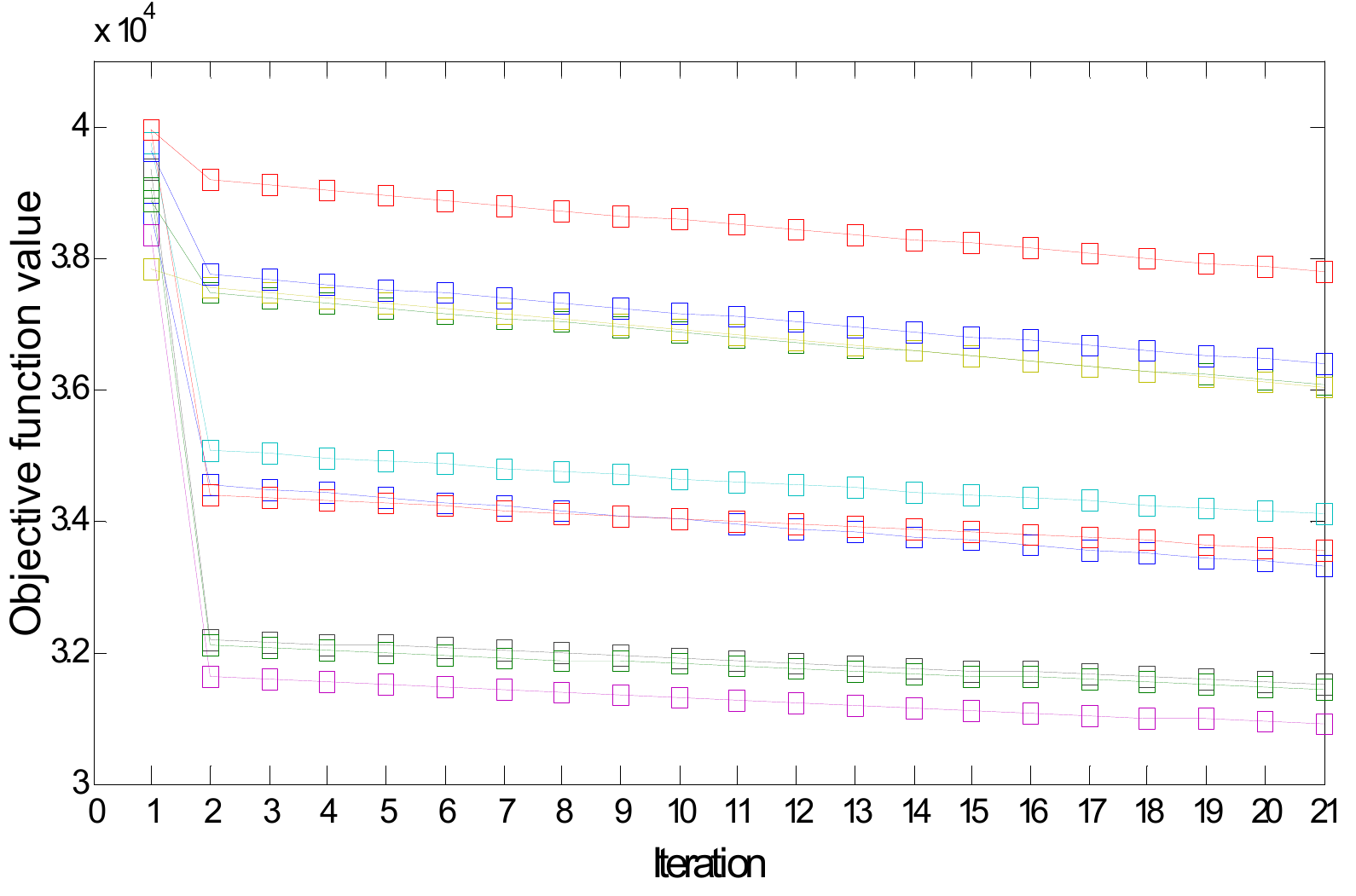}
\end{center}
   \caption{Convergence characteristics of the optimization \mbox{algorithm} of the proposed CERML-ES in the task of V2S face recognition on COX. Here, the 10 lines indicate the results of the 10 random V1-S testings on COX. The value '1' in x-axis is the case of the initialization.}
\label{fig_con_v2s}
\end{figure}

\subsubsection{Results and Analysis}

We summarize the V2S/S2V face recognition results on the PaSC and COX databases in Tab.\ref{tab_pasc_cox_v2s_s2v}, where each reported result on COX is the mean rank-1 identification rate over ten random runs of testing, while the presented results on PaSC are rank-1 verification rates at a false acceptance rate (FAR) = 0.01. In addition, we present comparisons in terms of running time in the supplementary material.



As shown in Tab.\ref{tab_pasc_cox_v2s_s2v}, besides to comparing the metric learning methods on gray features (/deep features), we also study the state-of-the-art deep learning method VGGDeepFace \cite{parkhi2015deep}. From the results, we can achieve some observations. Particularly, on PaSC, the homogeneous (Euclidean) metric learning methods generally outperform the existing heterogeneous metric learning methods. Nevertheless, by designing more sophisticated heterogeneous metric learning, our proposed CERML can achieve the results comparable with the state-of-the-art on this database. On COX, in most of tests, the exiting heterogeneous metric learning methods KCCA and KGMA coupled with our proposed heterogeneous data model are comparable with state-of-the-art homogeneous metric learning. In contrast, in the two tests of V3-S and S-V3, since the videos typically record more frontal face images with little variations, KCCA and KGMA cannot take advantage of the variation modeling on videos and thus are outperformed by the competing Euclidean metric learning approaches. In spite of this challenge, the proposed CERML achieves state-of-the-art results on all the COX tests. Especially, coupling with the deep features, the improvements of CERML over the existing state-of-the-art methods LMNN and KGMA are around 9\% and 7\%, respectively, on average. By comparing the deep learning method VGGDeepFace on COX and PaSC, we find that the CERML consistently makes a certain level of improvements (about 10\% on average) when using deep features. This verifies that our proposed first- and second-order pooling scheme can improve V2S/S2V face recognition.

Besides, we also present the performances of our CERML with different video models working on PaSC (deep feature) and COX (gray feature) in Fig.\ref{Fig5}. It can be found that jointly exploiting the first-order pooling (i.e., mean) and the second-order pooling (i.e., linear subspace, affine subspace, covariance matrix) significantly beats their separate exploitation, which was investigated in our prior work \cite{huang2014lerm}, in most of cases. This further validates the effectiveness of the extended CERML method.

In Fig.\ref{fig_con_v2s} we also report the convergence characteristics of the optimization \mbox{algorithm} of the proposed CERML-ES for the task of V2S/S2V face recognition on COX. As can be seen from Fig.\ref{fig_con_v2s}, the objective function Eq.\ref{Eq32} of the proposed CERML is able to converge to a stable and desirable solution after about 20 iterations. This mainly attributes to the well-designed initialization strategy (i.e., an initialization with the well-studied Fisher criterion) and the alternate updating on the parameter transformation matrices. In other words, adopting a good initialization strategy and regularizing the change in the transformation matrices will help to avoid the failure of convergence.


\subsection{V2V Face Recognition}

Video-to-Video (V2V) face recognition queries a video sequence against a set of target videos. In order to evaluate the proposed method working on such face recognition scenario, we use four publicly available video face databases, i.e., YTC \cite{kim2008ytc}, YTF \cite{wolf2011youtube}, PaSC \cite{beveridge2013challenge} and COX \cite{huang2015cox}.

\subsubsection{Dataset Setting}

For the YTC dataset, we resized each face image to a $20 \times 20$ image as was also done in \cite{wang2012covariance,lu2013image}, and pre-processed the resulting image with histogram equalization to eliminate lighting effects. As done in the last evaluation, we extract both of gray-level features and deeply learned features with VGGDeepFace \cite{parkhi2015deep} from each face image. Following the prior works \cite{wang2009manifold, wang2012covariance,lu2013image,hayat2015deep}, we conduct ten-fold cross validation experiments, i.e., 10 randomly selected gallery/probe combinations. In each fold, one person has 3 randomly chosen videos for the gallery and 6 for probes.

By employing the provided data of the YTF, we directly crop and normalize the face images from videos, and extract both gray features ande deep features from the normalized face images. For this database, we follow the standard evaluation protocol \cite{wolf2011youtube} to perform standard, ten-fold, cross validation, V2V face identification tests. Specifically, we utilize the officially provided 5,000 video pairs, which are equally divided into 10 folds. Each fold contains 250 intra-personal pairs and 250 inter-personal pairs.


For the PaSC dataset, based on the metadata provided in \cite{beveridge2013challenge}, we crop each face image to a color image of size $224 \times 224$. Similar to the V2S evaluation on PaSC, the work \cite{parkhi2015deep} is employed to extract deep face features on the normalized face images. For the V2V evaluation, there are two video face verification tests: control-to-control and handheld-to-handheld experiments. In both of them, the target and query sigsets contain the same set of videos. The task is to verify a claimed \mbox{identity} in the query video by comparing with the associated target video. Since the same 1,401 control/handheld videos serve as both the target and query sets, the `same video' comparisons are excluded.

The COX database, for V2V identification evaluations, has 3 videos per subject respectively from 3 camcorders. Therefore, they can generate 6 experiments. The 10 random partitions of the 300/700 subjects are designed for training and testing. In the V2V face recognition testing, each video face frame is normalized to a grayscale image. Similar to its V2S/S2V evaluation, we not only extract gray features, but also use deep features with the deep learning method of VGGDeepFace \cite{parkhi2015deep}.

\subsubsection{Method Setting}

In this evaluation, we compare the proposed CERML with three state-the-art deep learning methods being DeepFace \cite{taigman2014deepface}, FaceNet \cite{schroff2015facenet}, VGGDeepFace \cite{parkhi2015deep} and DisDeepFace \cite{wen2016discriminative}. Besides, we also study three groups of state-of-the-art Riemannian metric learning methods. The first category of methods first model videos with linear subspaces lying on a Grassmann manifold and then learn a Grassmannian metric for comparing two linear subspaces. The second group of methods first represents videos with affine subspaces residing on an Affine Grassmann manifold and then compute/learn the Riemannian metric for matching two affine subspaces. The third kind of methods first employs SPD matrices to represent videos and learns a Riemannian metric on the SPD manifold for comparing SPD data.

\begin{enumerate}
   \item Grassmannian metric learning methods:

     Discriminative Canonical Correlations (DCC) \cite{kim2007discriminative}, Grassmann Discriminant Analysis (GDA) \cite{hamm2008gda}, Grassmannian Graph-Embedding Discriminant Analysis (GGDA) \cite{harandi2011ggda}, Projection Metric Learning (PML) \cite{huang2015projection};

  \item Affine Grassmannian computing/metric learning methods:

     Affine Hull based Image Set Distance (AHISD) \cite{cevikalp2010face}, Convex Hull based Image Set Distance (CHISD) \cite{cevikalp2010face}, Set-to-Set Distance Metric Learning (SSDML) \cite{zhu2013p2s};

  \item SPD Riemannian metric learning methods:

    Localized Multi-Kernel Metric Learning (LMKML) \cite{lu2013image}, Covariance Discriminative Learning (CDL) \cite{wang2012covariance}.

\end{enumerate}

In order to achieve a fair comparison, the key parameters of each method are empirically tuned according to the recommendations in the original works. For MSM/AHISD, the first canonical correlation or leading component is exploited when comparing \mbox{two} subspaces. For the first group of methods DCC/GDA/GGDA, the dimensionality of the resulting discriminant subspace is tuned from 1 to 10. For GDA/GGDA, the final dimensionality is set to $c-1$ ($c$ is the number of face classes in training). For AHISD, the leading component is exploited when comparing two affine subspaces. For SSDML, its key parameters are tuned and empirically set as: $\lambda_1=0.001, \lambda_2=0.5$, the numbers of positive and negative pairs per sample are 10 and 20 respectively. In GGDA, the combination parameter $\beta$ is tuned around the value of 100. For LMKML, the widths of Gaussian kernels are tuned around the mean distance. For our CERML, similar to the V2S/S2V evaluation, the regularization parameters are set as $\lambda_1=0.01, \lambda_2=0.1$, the neighborhood number $k_1 = 1, k_2 =20$, the kernel widths $\sigma$s are equal to the mean distances on the training multi-view data, and the number of iterations is set to $20$.

\begin{table}[t]
\linespread{1.2}
\caption{V2V face recognition results (\%) on YTC and YTF using gray/deep features. Here EG, EA and ES are the Euclidean-to-Grassmannian, Euclidean-to-AffineGrassmannian and Euclidean-to-SPD matchings.}
\footnotesize
\begin{center}
\begin{tabular}{|m{2.5cm}<{\centering}|m{2cm}<{\centering}||m{2cm}<{\centering}|}
\hline
Methods & YTC \emph{(gray/deep)} & YTF \emph{(gray/deep)} \\
\hline\hline
DeepFace \cite{taigman2014deepface} & --/-- & --/91.4 \\
FaceNet \cite{schroff2015facenet} & --/-- & --/\textbf{95.1} \\
VGGDeepFace \cite{parkhi2015deep} & --/83.74 & --/91.78 \\
DisDeepFace \cite{wen2016discriminative} & --/-- & --/94.9 \\
\hline
DCC \cite{kim2007discriminative} & 68.85/86.15 & 68.28/92.52\\
GDA \cite{hamm2008gda} & 65.02/86.44 & 67.00/89.20\\
GGDA \cite{harandi2011ggda} & 66.56/86.54 & 66.56/91.60 \\
PML \cite{huang2015projection} & 66.69/86.98 & 67.30/92.58 \\
\hline
AHISD \cite{cevikalp2010face} & 66.37/81.49 & 64.80/90.74\\
CHISD \cite{cevikalp2010face} & 66.62/74.88 & 66.30/90.00\\
SSDML \cite{zhu2013p2s} & 68.85/85.59 & 65.38/88.26\\
\hline
CDL \cite{wang2012covariance} & 69.72/85.98 & 64.94/90.65\\
LMKML \cite{lu2013image} & 68.13/85.16  & 64.39/89.53\\
\hline
CERML-EG & 68.08/87.62 & 69.42/93.36\\
CERML-EA & 69.57/88.01 & 68.89/94.06\\
CERML-ES & 72.38/\textbf{88.51}  & 68.36/93.44  \\
\hline
\end{tabular}
\end{center}
\label{tab_ytc_ytf_v2v}
\end{table}

\begin{table*}[t]
\linespread{1.2}
\caption{V2V face recognition results (\%) on PaSC and COX using gray/deep features. Here EG, EA and ES indicate Euclidean-to-Grassmannian, Euclidean-to-AffineGrassmannian and Euclidean-to-SPD matching, respectively. Con and han represent the control and handheld settings.}
\footnotesize
\begin{center}
\begin{tabular}{|m{2.4cm}<{\centering}|m{1.45cm}<{\centering}|m{1.45cm}<{\centering}||m{1.4cm}<{\centering}|m{1.4cm}<{\centering}
|m{1.4cm}<{\centering}|m{1.4cm}<{\centering}|m{1.4cm}<{\centering}|m{1.4cm}<{\centering}|}\hline
\multirow{2}{*}{Methods} & \multicolumn{2}{c||}{PaSC \emph{(deep)}} & \multicolumn{6}{c|}{COX \emph{(gray/deep)}}\\
\cline{2-9} & V(con)-V(con) & V(han)-V(han) & V2-V1 & V3-V1 & V3-V2 &V1-V2 &V1-V3 & V2-V3 \\
\hline\hline
VGGDeepFace\cite{parkhi2015deep} & 78.82 & 68.24 & --/92.24 & --/87.54 & --/91.63 & --/92.34 & --/92.47 & --/95.96  \\
\hline
DCC \cite{kim2007discriminative} & 75.83 & 67.04 & 62.53/95.86 & 66.10/95.57 & 50.56/93.00 & 56.09/94.29 & 53.84/96.86  & 45.19/96.29\\
GDA \cite{hamm2008gda} & 71.38 & 67.49 & 68.61/95.11 & 77.70/95.87 & 71.59/95.16 & 65.93/94.41 & 76.11/96.10  & 74.83/96.26\\
GGDA \cite{harandi2011ggda} & 66.71 & 68.41 & 70.80/95.81 & 76.23/96.30 & 71.99/95.61 & 69.17/95.34 & 76.77/96.66  & 77.43/96.61\\
PML \cite{huang2015projection} & 73.45 & 68.32 & 71.27/95.57 & 78.91/95.43 & 73.24/93.29 & 64.62/95.86 & 78.26/97.13 & 78.15/97.00 \\
\hline
AHISD \cite{cevikalp2010face} & 53.93 & 45.11 & 53.03/92.85 & 36.13/95.57 & 17.50/95.43 & 43.51/93.57 & 34.99/92.43  & 18.80/94.71\\
CHISD \cite{cevikalp2010face} & 60.54 & 47.00 & 56.90/92.43 & 30.13/94.57 & 15.03/93.43 & 44.36/91.57 & 26.40/94.06  & 13.69/94.57\\
SSDML \cite{zhu2013p2s} &  65.32  & 56.23 & 60.13/77.43  & 53.14/90.71 & 28.73/89.57 & 47.91/70.43 & 44.42/85.43 & 27.34/87.57\\
\hline
CDL \cite{wang2012covariance} & 72.69 & 65.44 & 78.43/95.53 &  85.31/97.61 & 79.71/96.40 & 75.56/95.96 & 85.84/97.33  & 81.87/96.51 \\
LMKML \cite{lu2013image} & 70.41  & 66.15 & 56.14/94.73  & 44.26/95.16 & 33.14/96.37 & 55.37/93.81 & 39.83/96.12 & 29.54/96.28\\
\hline
CERML-EG & \textbf{80.11} & \textbf{77.37} & 87.59/97.77 & 92.41/98.07 & 88.54/97.39 & 83.21/97.66 & 92.09/\textbf{98.59} & 91.16/\textbf{97.77} \\
CERML-EA & 77.71 & 75.03 & 87.14/\textbf{98.26} & 91.94/\textbf{98.33} & 88.30/\textbf{97.60} & 82.81/\textbf{97.97} & 92.03/98.49 & 91.16/97.66\\
CERML-ES & 79.92 & 76.92 & 90.31/98.17 & 94.83/98.27 & 91.51/97.46 & 87.06/97.71 & 95.13/98.31 & 93.89/97.64\\
\hline
\end{tabular}
\end{center}
\label{tab_pasc_cox_v2v}
\end{table*}

\begin{figure}[t]
\begin{center}
\includegraphics[width=0.9\linewidth]{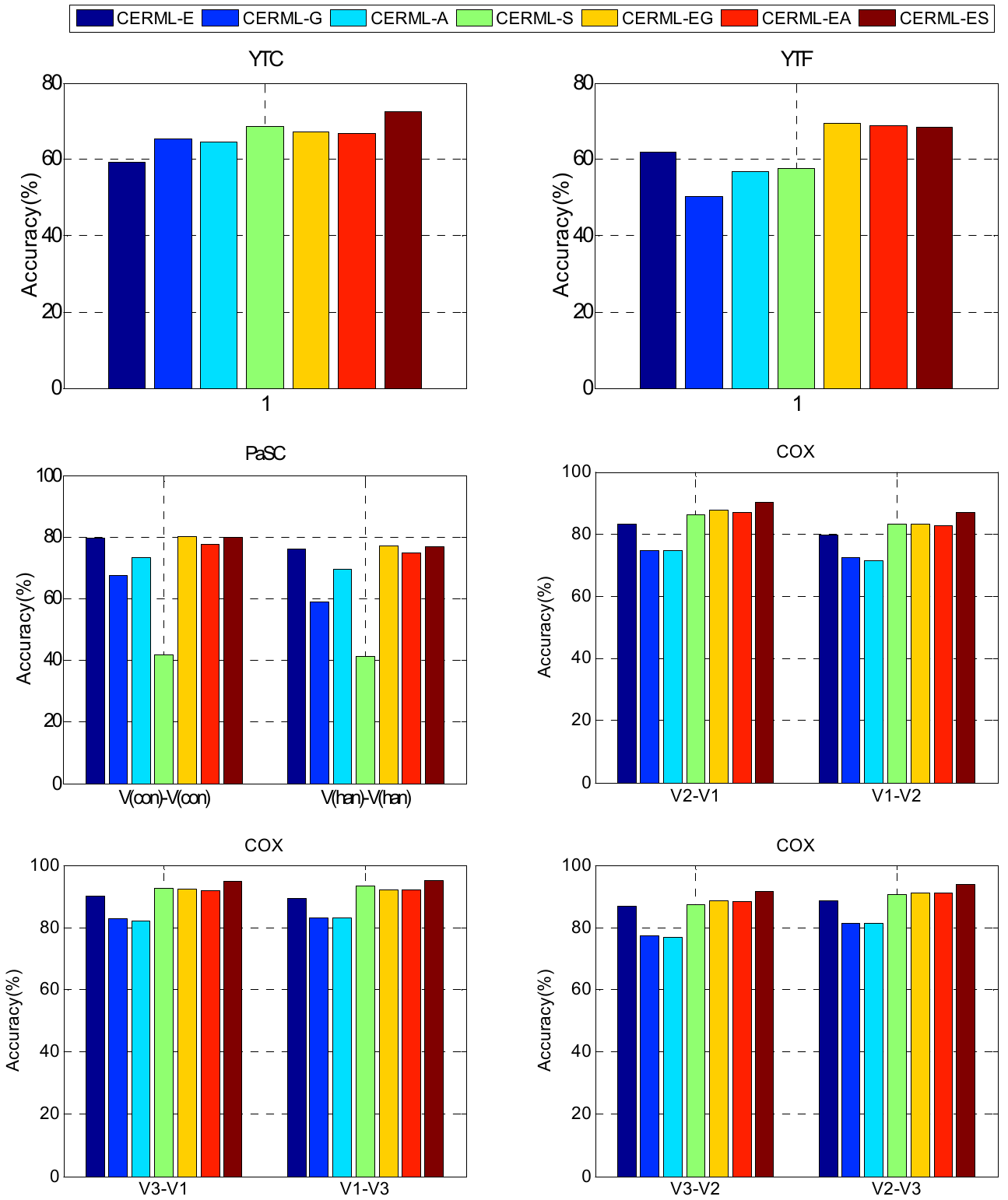}

\end{center}
\caption{V2V face recognition results (\%) of the proposed CEMRL dealing with different representations of videos using gray features for YTC, YTF, COX and deep features for PaSC. Here, CERML-E, CERML-G, CERML-A, CERML-S, CERML-EG, CERML-EA, CERML-ES respectively indicate videos are represented by mean, subspace, affine subspace, SPD matrix, mean+subspace, mean+affine subspace, mean+SPD matrix.}
\label{Fig6}
\end{figure}

\begin{figure}[t]
\begin{center}
   \includegraphics[width=0.8\linewidth]{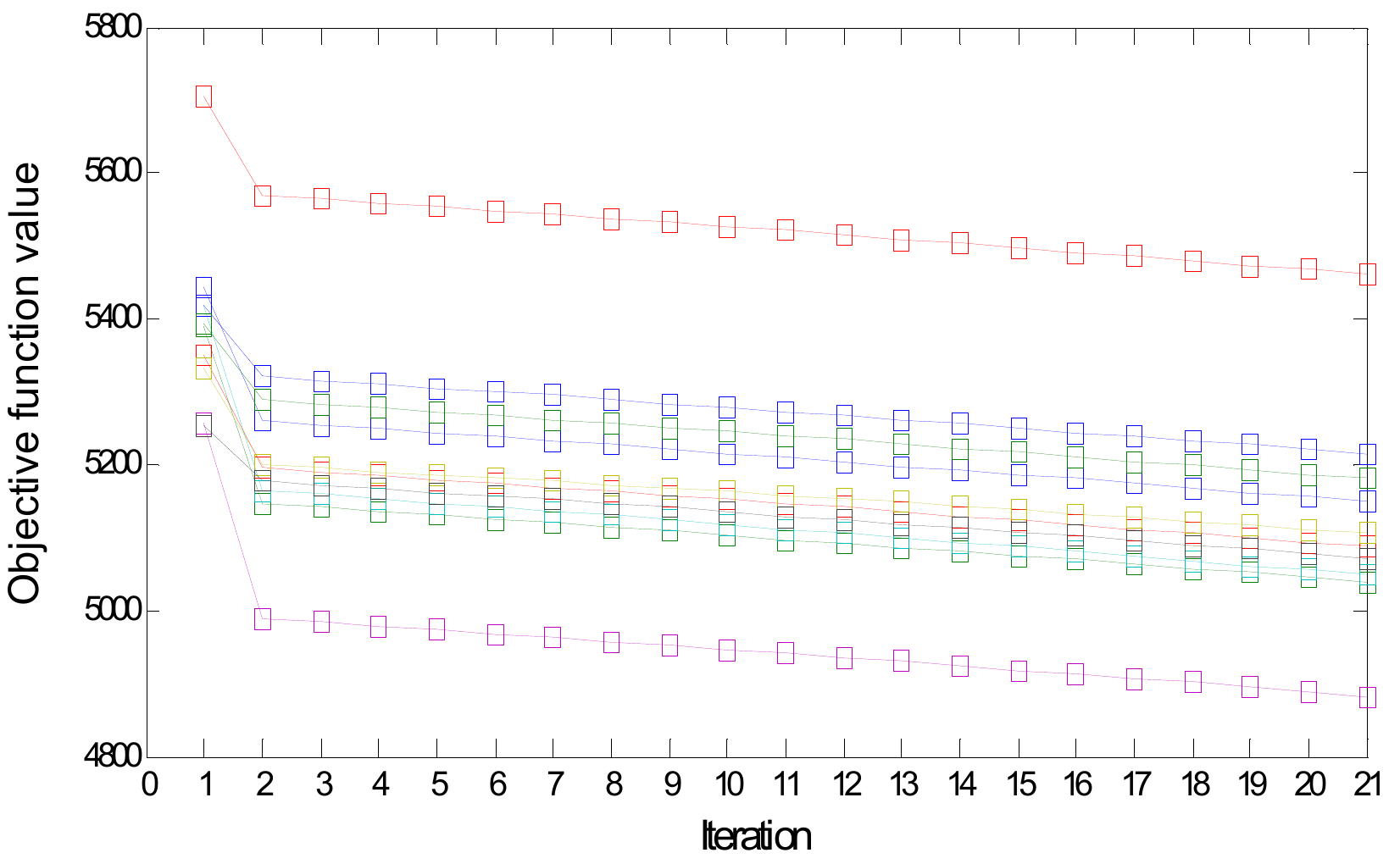}
\end{center}
   \caption{Convergence behaviors of the optimization \mbox{algorithm} of the proposed CERML-ES in the task of V2V face recognition for COX. Here, the 10 lines respectively indicate the results of the 10 random V2-V1 tests on COX. The value '1' along the x-axis is the case of the initialization.}
\label{fig_con_v2v}
\end{figure}

\subsubsection{Results and Analysis}

The V2V face recognition tests on the four video face databases are summarized in Tab.\ref{tab_ytc_ytf_v2v} and Tab.\ref{tab_pasc_cox_v2v}, where each accuracy on YTC and COX is the mean rank-1 identification rate over ten random runs of testing, while the results on YTF and PaSC are mean verification rates and rank-1 verification rates at FAR = 0.01, respectively. Besides, the running time of the competing methods are also presented in the supplementary material.

From the results on the internet video face databases YTC and YTF in Tab.\ref{tab_ytc_ytf_v2v}, it can be observed that the three existing types of Riemannian metric learning achieve comparable V2V face recognition accuracies. This demonstrates that the Riemannian representations, i.e., linear subspace and SPD matrix, employed by such Riemannian metric learning methods typically encode approximately the same information in terms of variations of videos for video-based face recognition. In contrast, the proposed CERML simultaneously exploits the mean and the variation, resulting in more robust face recognition. Hence, CERML generally outperform all the competing methods on the two \mbox{databases}. In addition, by studying the behavior of the state-of-the deep learning methods, we observe that our CERML can be comparable with or even surpass them.

The evaluations on the two surveillance-like video face databases PaSC and COX are reported in Tab.\ref{tab_pasc_cox_v2v}. Approximately the same conclusions can be drawn from the results on the two datasets. On PaSC, the three affine Grassmannian learning methods are generally outperformed by the other two kinds of Riemannian metric learning methods. This may be because AHISD and CHISD exploit the appropriate distance metric between Riemannian representations without employing the label information of the training data. Besides, SSDML just treats each video with an affine combination of frames without considering its Riemannian structure, and thus yields worse performances on PaSC. By fusing the mean and the variation information on videos, our CERML-EG performs the best with improvements of about 4\% and 10\% over the state-of-the-art method DCC in the two tests on PaSC, respectively. On COX, lower performances are also achieved by the three affine Grassmannian metric learning methods for the same reasons mentioned above. Compared with the existing Riemannian metric learning methods, the proposed CERML can achieve the state-of-the-art in all the V2V face recognition tests on COX. Specifically, whenever using gray features or deep features, CERML always reaches some improvements over the existing state-of-the-art metric learning methods. Furthermore, we find that the proposed CERML with deep features has an average gain of 5\% over the state-of-the-art deep learning method VGGDeepFace on the PaSC and COX databases.

In addition, we also report the performances of our CERML with different video models working on YTC, YTF, PaSC and COX in Fig.\ref{Fig6}. As can be seen, simultaneously exploiting the mean appearance and pattern variance typically performs better than exploiting them separately. This further demonstrates the effectiveness of fusing Euclidean and Riemannian representations in the proposed metric learning scheme in our CERML.

In the end, we present the convergence behavior of the optimization \mbox{algorithm} of the proposed CERML-ES in the scenario of V2V face recognition on COX (see Fig.\ref{fig_con_v2v}). As seen, the objective function Eq.\ref{Eq38} of the algorithm can converge to a desirable value after tens of iterations.

\section{Conclusion and Future Work}

In this paper, we have introduced a novel heterogeneous metric learning framework that works across a Euclidean space and a Riemannian manifold to match/fuse Euclidean and Riemannian representations of videos. The proposed method offers a unified framework for three typical tasks of video-based face recognition. Our extensive experimental evaluations have clearly shown that the new technique can achieve the state-of-the-art on four challenging video face datasets in the three video-based face recognition tasks.

Our work contributes to learn a cross-view metric from a Euclidean space to a Riemannian manifold, by exploiting the Riemannian metric learning scheme of kernel embedding which introduces several intrinsic drawbacks (e.g., undesirable scalability) of traditional kernel learning methods as well. For future work, studying how to improve the proposed framework with other more effective Riemannian metric learning schemes such as manifold-to-manifold embedding would be very interesting.


\ifCLASSOPTIONcompsoc
  \section*{Acknowledgments}

\else
  \section*{Acknowledgment}
\fi

This work has been carried out mainly at the Institute of Computing Technology (ICT), Chinese Academy of Sciences (CAS). It is partially supported by 973 Program under contract No. 2015CB351802, Natural Science Foundation of China under contracts Nos. 61390511, 61173065, 61222211, and 61379083.

\begin{IEEEbiographynophoto}
{Zhiwu Huang}
received the B.S. degree in Computer Science and Technology from Huaqiao \mbox{University}, Quanzhou, Fujian, China, in 2007, and the M.S. degree in Computer Software and Theory from Xiamen University, Xiamen, Fujian, China, in 2010, and the Ph.D. degree in Computer Science and Technology from the Institute of Computing Technology (ICT), Chinese Academy of Sciences (CAS), Beijing, China, in 2015. He is currently a postdoctoral researcher with the Computer Vision Laboratory, Swiss Federal Institute of Technology (ETH) in Zurich, Switzerland, since September 2015. His research interests include computer vision, Riemannian computing, metric learning and deep learning.
\end{IEEEbiographynophoto}

\begin{IEEEbiographynophoto}
{Ruiping Wang}
received the B.S. degree in applied mathematics from Beijing Jiaotong University, Beijing, China, in 2003, and the Ph.D. degree in computer science from the Institute of Computing Technology (ICT), Chinese Academy of Sciences (CAS), Beijing, in 2010. He was a postdoctoral researcher with the Department of Automation, Tsinghua University, Beijing, from July 2010 to June 2012. He also spent one year working as a Research Associate with the Computer Vision Laboratory, Institute for Advanced Computer Studies, at the University of Maryland, College Park, from November 2010 to October 2011. He has been with the faculty of the Institute of Computing Technology, Chinese Academy of Sciences, since July 2012, where he is currently an Associate \mbox{Professor}. His interests include computer vision, pattern recognition, and machine learning.
\end{IEEEbiographynophoto}

\begin{IEEEbiographynophoto}{Shiguang Shan}
received M.S. degree in computer science from the Harbin Institute of Technology, Harbin, China, in 1999, and Ph.D. degree in computer science from the Institute of Computing Technology (ICT), Chinese Academy of Sciences (CAS), Beijing, China, in 2004. He joined ICT, CAS in 2002 and has been a Professor since 2010. He is now the Deputy Director of the Key Lab of Intelligent Information Processing of CAS. His research interests cover computer vision, pattern recognition, and machine learning. He especially focuses on face recognition related research topics. He has published more than 200 papers in refereed journals and proceedings in the areas of Computer Vision and Pattern Recognition. He has served as Area Chair for many international conferences including ICCV'11, ICPR'12, ACCV'12, FG'13, ICPR'14, and ICASSP'14. He served as workshop co-chair of ACCV14, and website co-chair of ICCV'15. He is Associate Editors of IEEE Trans. on Image Processing, Neurocomputing, and EURASIP Journal of Image and Video Processing. He received the China's State Scientific and Technological Progress Awards in 2005 for his work on face recognition technologies.
\end{IEEEbiographynophoto}

\begin{IEEEbiographynophoto}	
{Luc Van Gool}
received a degree in electro-mechanical engineering at the Katholieke Universiteit Leuven in 1981. Currently, he is a full professor for Computer Vision at the ETH in Zurich and the Katholieke Universiteit Leuven in Belgium. He leads research and teaches at both places. He has authored over 300 papers in his field. He has been a program committee member of several, major computer vision conferences (e.g. Program Chair ICCV'05, Beijing, and General Chair of ICCV'11, Barcelona, and of ECCV'14, Zurich). His main interests include 3D reconstruction and modeling, object recognition, and tracking and gesture analysis. He received several Best Paper awards (eg. David Marr Prize '98, Best Paper CVPR'07, Tsuji Outstanding Paper Award ACCV'09, Best Vision Paper ICRA'09). In 2015 he received the 5-yearly Excellence Prize for Applied Sciences from the Flemish Institute for Scientific Research (FWO). He is a co-founder of about 10 spin-off companies.
\end{IEEEbiographynophoto}

\begin{IEEEbiographynophoto}
 {Xilin Chen} 
received the B.S., M.S., and Ph.D. degrees in computer science from the Harbin Institute of Technology, Harbin, China, in 1988, 1991, and 1994, respectively. He was a professor with the Harbin Institute of Technology from 1999 to 2005. He was a visiting scholar with Carnegie Mellon University, Pittsburgh, PA, from 2001 to 2004. He has been a professor with the Institute of Computing Technology, Chinese Academy of Sciences (CAS), since August 2004. He is the Director of the Key Laboratory of Intelligent Information Processing, CAS. He has published one book and over 200 papers in refereed journals and proceedings in the areas of computer vision, pattern recognition, image processing, and multimodal interfaces. He is a leading editor of the Journal of Computer Science and Technology, and an associate editor in chief of the Chinese Journal of Computers. He served as an Organizing Committee / Program Committee Member for more than 50 conferences. He is a recipient of several awards, including the China's State Scientific and Technological Progress Award in 2000, 2003, 2005, and 2012 for his research work. He is a Fellow of the China Computer Federation (CCF).
\end{IEEEbiographynophoto}




\newpage

\section{Supplementary Material}
This Supplementary Material provides extra details on the following:
\begin{itemize}
	\item APPENDIX A: Detail of the initialization of the proposed CERML.
	\item APPENDIX B: Comparison of using linear kernels and RBF kernels.
	\item APPENDIX C: Benefit of exploiting the cross-view (Euclidean-to-Riemannian) kernels.
	\item APPENDIX D: Running time of the proposed CERML and the comparative methods.
\end{itemize}

\appendices

\section{Detail of the initialization of the proposed CERML}

In this section, we present more details of the initialization of our optimization algorithm. Specifically, we define the within-class and between-class templates for $\bm{A}$, $\bm{A}_x$ and $\bm{A}_{y}$  in Eq.11, Eq.12 and Eq.13 respectively in the main paper as:
\begin{equation}
\bm{A}^w(i,j) = \begin{cases}
1, &\text{if $l^x_i = l^y_j$},\\
0, &\text{if $l^x_i \neq l^y_j$}.
\end{cases}
\label{Eq113}
\end{equation}
\begin{equation}
\bm{A}^b(i,j) = \begin{cases}
0, &\text{if $l^x_i = l^y_j$},\\
1, &\text{if $l^x_i \neq l^y_j$}.
\end{cases}
\label{Eq114}
\end{equation}
\begin{equation}
\bm{A}^w_x(i,j) =
\begin{cases}
d_{ij}, &\text{if $l^x_i = l^x_j$ and $k_1(i,j)$},\\
0, &\text{else}.
\end{cases}
\label{Eq115}
\end{equation}
\begin{equation}
\bm{A}^b_x(i,j) =
\begin{cases}
d_{ij}, &\text{if $l^x_i \neq l^x_j$ and $k_2(i,j)$},\\
0, &\text{else}.
\end{cases}
\label{Eq116}
\end{equation}
\begin{equation}
\bm{A}^w_y(i,j) =
\begin{cases}
d_{ij}, &\text{if $l^y_i = l^y_j$ and $k_1(i,j)$},\\
0, &\text{else}.
\end{cases}
\label{Eq117}
\end{equation}
\begin{equation}
\bm{A}^b_y(i,j) =
\begin{cases}
d_{ij}, &\text{if $l^y_i \neq l^y_j$ and $k_2(i,j)$},\\
0, &\text{else}.
\end{cases}
\label{Eq118}
\end{equation}

By substituting Eq.\ref{Eq113} and Eq.\ref{Eq114} into Eq.11 (in the main paper), the within-class $D^w(\bm{W}_x,\bm{W}_y)$ and between-class templates $D^b(\bm{W}_x,\bm{W}_y)$ for $D(\bm{W}_x,\bm{W}_y)$ can be achieved as��
\begin{equation}
\begin{aligned}
D^w(\bm{W}_x,\bm{W}_y) & = \frac{1}{2} (\bm{W}_x^T\bm{K}_{x}\bm{B}^{'w}_x\bm{K}_{x}^T\bm{W}_x +\bm{W}_y^T\bm{K}_{y}\bm{B}^{'w}_y\bm{K}_{y}^T\bm{W}_y\\&-2\bm{W}_x^T\bm{K}_{x}\bm{Z}^w\bm{K}_{y}\bm{W}_y^T).
\label{Eq119}
\end{aligned}
\end{equation}
\begin{equation}
\begin{aligned}
D^b(\bm{W}_x,\bm{W}_y) & = \frac{1}{2} (\bm{W}_x^T\bm{K}_{x}\bm{B}^{'b}_x\bm{K}_{x}^T\bm{W}_x +\bm{W}_y^T\bm{K}_{y}\bm{B}^{'b}_y\bm{K}_{y}^T\bm{W}_y\\&-2\bm{W}_x^T\bm{K}_{x}\bm{Z}^b\bm{K}_{y}\bm{W}_y^T).
\label{Eq120}
\end{aligned}
\end{equation}

Likewise, using Eq.\ref{Eq115}, Eq.\ref{Eq116} and Eq.12 (in the main paper) can achieve the within-class $G_x^w(\bm{W}_x)$ and between-class templates $G_x^b(\bm{W}_x)$ for $G_x(\bm{W}_x)$, while employing Eq.\ref{Eq117}, Eq.\ref{Eq118} and Eq.13 (in the main paper) can obtain the within-class $G_y^w(\bm{W}_y)$ and the between-class templates $G_y^w(\bm{W}_y)$ for $G_y(\bm{W}_y)$:
\begin{equation}
\begin{aligned}
G^w_x(\bm{W}_x) &= \bm{W}_x^T\bm{K}_{x}\bm{B}^w_x\bm{K}_{x}^T\bm{W}_x-\bm{W}_x^T\bm{K}_{x}\bm{Z}^w_x\bm{K}_{x}^T\bm{W}_x\\
& = \bm{W}_x^T\bm{K}_{x}\bm{L}^w_x\bm{K}_{x}^T\bm{W}_x.
\label{Eq121}
\end{aligned}
\end{equation}
\begin{equation}
\begin{aligned}
G^b_x(\bm{W}_x) &= \bm{W}_x^T\bm{K}_{x}\bm{B}^b_x\bm{K}_{x}^T\bm{W}_x-\bm{W}_x^T\bm{K}_{x}\bm{Z}^b_x\bm{K}_{x}^T\bm{W}_x\\
& = \bm{W}_x^T\bm{K}_{x}\bm{L}^b_x\bm{K}_{x}^T\bm{W}_x.
\label{Eq122}
\end{aligned}
\end{equation}
\begin{equation}
\begin{aligned}
G^w_y(\bm{W}_y) &= \bm{W}_y^T\bm{K}_{y}\bm{B}^w_y\bm{K}_{y}^T\bm{W}_y-\bm{W}_y^T\bm{K}_{y}\bm{Z}^w_y\bm{K}_{y}^T\bm{W}_y\\
& = \bm{W}_y^T\bm{K}_{y}\bm{L}^w_y\bm{K}_{y}^T\bm{W}_y.
\label{Eq123}
\end{aligned}
\end{equation}
\begin{equation}
\begin{aligned}
G^b_y(\bm{W}_y) &= \bm{W}_y^T\bm{K}_{y}\bm{B}^b_y\bm{K}_{y}^T\bm{W}_y-\bm{W}_y^T\bm{K}_{y}\bm{Z}^b_y\bm{K}_{y}^T\bm{W}_y\\
& = \bm{W}_y^T\bm{K}_{y}\bm{L}^b_y\bm{K}_{y}^T\bm{W}_y.
\label{Eq124}
\end{aligned}
\end{equation}

Then we can initialize $\bm{W}_x$ and $\bm{W}_{y}$ by maximizing the sum of between-class templates while minimizing the sum of within-class templates as the Fisher criterion of the traditional Fisher Discriminant Analysis (FDA) \cite{fisher1936use}:
\begin{equation}
\begin{aligned}
\mathop{\max}_{\bm{W}_x,\bm{W}_y} & \{\bm{D}^b(\bm{W}_x,\bm{W}_y) + \lambda_1 G^b(\bm{W}_x,\bm{W}_y)\},\\
s.t. \quad & \bm{D}^w(\bm{W}_x,\bm{W}_y) + \lambda_1 G^w(\bm{W}_x,\bm{W}_y) = 1.
\label{Eq125}
\end{aligned}
\end{equation}
where $G^b(\bm{W}_x, \bm{W}_y)=G^b_x(\bm{W}_x, \bm{W}_y)+G^b_y(\bm{W}_x, \bm{W}_y)$�� $G^w(\bm{W}_x, \bm{W}_y)=G^w_x(\bm{W}_x, \bm{W}_y)+G^w_y(\bm{W}_x, \bm{W}_y)$. Substitute Eq.\ref{Eq119}-Eq.\ref{Eq124} into Eq.\ref{Eq125} to obtain��
\begin{equation}
\begin{aligned}
\max  & \begin{bmatrix}\bm{W}_x\\\bm{W}_y\\\end{bmatrix}^T
\begin{bmatrix}\bm{K}_{x}\bm{R}_x^{b}\bm{K}_{x}^T & -\bm{K}_{x}\bm{A}^b\bm{K}_{y}^T\\-\bm{K}_{y}({\bm{A}^b})^T \bm{K}_{x}^T & \bm{K}_y\bm{R}_y^{b}\bm{K}_{y}^T\\\end{bmatrix}
\begin{bmatrix}\bm{W}_x\\\bm{W}_y\\\end{bmatrix}\\
s.t. & \begin{bmatrix}\bm{W}_x\\\bm{W}_y\\\end{bmatrix}^T
\begin{bmatrix}\bm{K}_{x}\bm{R}_x^{w}\bm{K}_{x}^T & -\bm{K}_{x}\bm{A}^w\bm{K}_{y}^T\\-\bm{K}_{y}({\bm{A}^w})^T \bm{K}_{x}^T & \bm{K}_y\bm{R}_y^{w}\bm{K}_{y}^T\\\end{bmatrix}
\begin{bmatrix}\bm{W}_x\\\bm{W}_y\\\end{bmatrix} = 1.
\label{Eq126}
\end{aligned}
\end{equation}
where $\bm{R}_x^{b}=\bm{B}_x^b+2\lambda_1 \bm{L}_x^b$, $\bm{R}_y^{b}=\bm{B}_y^b+2\lambda_1 \bm{L}_y^b$, $\bm{R}_x^{w}=\bm{B}_x^w+2\lambda_1 \bm{L}_x^w$�� $\bm{R}_y^{w}=\bm{B}_y^w+2\lambda_1 \bm{L}_y^w$. Equivalently, the optimization function can be further simplified to:
\begin{equation}
\begin{aligned}
&\max \bm{W}^T\bm{M}^b\bm{W}, \quad s.t.\bm{W}^T\bm{M}^w\bm{W}= 1. \\
&\Rightarrow \bm{M}^b\bm{W} = \lambda \bm{M}^w\bm{W}.
\label{Eq127}
\end{aligned}
\end{equation}
where matrix $\bm{M}^b=\begin{bmatrix}\bm{K}_{x}\bm{R}_x^{b}\bm{K}_{x}^T & -\bm{K}_{x}\bm{Z}^b\bm{K}_{y}^T\\-\bm{K}_{y}({\bm{Z}^b})^T \bm{K}_{x}^T & \bm{yR}_y^{b}\bm{K}_{y}^T\\\end{bmatrix},\bm{M}^w=\begin{bmatrix}\bm{K}_{x}\bm{R}_x^{w}\bm{K}_{x}^T & -\bm{K}_{x}\bm{Z}^w\bm{K}_{y}^T\\-\bm{K}_{y}({\bm{Z}^w})^T \bm{K}_{x}^T & \bm{yR}_y^{w}\bm{K}_{y}^T\\\end{bmatrix}$, $\bm{W}^T=[\bm{W}_x^T, \bm{W}_y^T]$. Obviously, the final objective function is a standard generalized eigenvalue problem that can be solved using any eigensolver.

\begin{table*}[t]
	\linespread{1.2}
	\caption{V2S/S2V face recognition results (\%) of the proposed CERML-EG/EA/ES with linear and RBF kernel settings and using deep features on the COX database. Here EG, EA and ES are the Euclidean-to-Grassmannian, Euclidean-to-AffineGrassmannian and Euclidean-to-SPD matchings.}
	\footnotesize
	\begin{center}
		\begin{tabular}{|m{3cm}<{\centering}|m{1.5cm}<{\centering}|m{1.5cm}<{\centering}|m{1.5cm}<{\centering}|m{1.5cm}<{\centering}|m{1.5cm}<{\centering}|m{1.5cm}<{\centering}|}
			\hline
			Methods & V1-S & V2-S & V3-S & S-V1 & S-V2 & S-V3 \\ 
			\hline\hline
			
			CERML-EG (linear) & 85.36 & 82.19 & 87.19 & 88.93 & 85.60 & 90.29\\
			CERML-EG (RBF) & 85.71 & 82.51 & 87.23 & 88.80 & 85.69 & 90.99\\
			\hline
			CERML-EA (linear) & 86.17 & 83.60 & 87.71 & 89.30 & 85.61 & 90.79\\
			CERML-EA (RBF) & 86.40 & 83.13 & 86.76 & 88.97 & 85.84 & 90.26\\
			\hline
			CERML-ES (linear) & 84.67 & 81.29 & 85.96 & 86.57 & 83.50 & 86.97\\
			CERML-ES (RBF) & 86.21 & 82.66 & 86.64 & 88.93 & 85.37 & 89.64\\
			\hline
		\end{tabular}
	\end{center}
	\label{tab_linear_rbf_v2s}
\end{table*}

\begin{table*}[t]
	\linespread{1.2}
	\caption{V2V face recognition results (\%) of the proposed CERML-EG/EA/ES with linear and RBF kernel settings and using deep features on the COX database. Here EG, EA and ES are the Euclidean-to-Grassmannian, Euclidean-to-AffineGrassmannian and Euclidean-to-SPD matchings.}
	\footnotesize
	\begin{center}
		\begin{tabular}{|m{3cm}<{\centering}|m{1.5cm}<{\centering}|m{1.5cm}<{\centering}|m{1.5cm}<{\centering}|m{1.5cm}<{\centering}|m{1.5cm}<{\centering}|m{1.5cm}<{\centering}|}
			\hline
			Methods & V2-V1 & V3-V1 & V3-V2 & V1-V2 & V1-V3 & V2-V3 \\ 
			\hline\hline
			
			CERML-EG (linear) & 97.84 & 98.10 & 97.40 & 97.63 & 98.57 & 97.81\\
			CERML-EG (RBF) & 98.77 & 98.07 & 97.39 & 97.66 & 98.59 & 97.77\\
			\hline
			CERML-EA (linear) & 98.03 & 98.09 & 97.54 & 97.73 & 98.63 & 97.90\\
			CERML-EA (RBF) & 98.76 & 98.33 & 97.60 & 97.97 & 98.49 & 97.66\\
			\hline
			CERML-ES (linear) & 98.39 & 98.33 & 98.77 & 98.04 & 98.71 & 97.93\\
			CERML-ES (RBF) & 98.71 & 98.27 & 97.46 & 97.71 & 98.31 & 97.64\\
			\hline
		\end{tabular}
	\end{center}
	\label{tab_linear_rbf_v2v}
\end{table*}

\section{Comparison of using linear kernels and RBF kernels}

As the RBF kernel is one of the most common kernels, we merely employ it as a special case for Eq.5 and Eq.6 in the main paper. In addition to it, there are many popular kernels such as linear kernel, which is able to be applied to our metric learning framework as well. In the material, we conduct some experiments to suggest using these two kernels can achieve approximately the same performances for the COX V2S/S2V and V2V evaluations as shown in Tab.\ref{tab_linear_rbf_v2s} and Tab.\ref{tab_linear_rbf_v2v}.

\section{Benefit of exploiting the cross-view (Euclidean-to-Riemannian) kernels}

For V2S/S2V face recognition, in addition to employing the single-view kernels for Euclidean and Riemannian representations, we also defined cross-view kernels to concatenate the single-view kernels. As a result, the final kernels coupled in our proposed CERML method is defined as: $\hat{\bm{K}}_x=[\bm{K}_x, \bm{K}_{xy}]$, $\hat{\bm{K}_{y}}=[\bm{K}_{y}, (\bm{K}^{'}_{xy})^T]$, where $\bm{K}_x, \bm{K}_y$ are the single-view kernels, $\bm{K}_{xy}$ is the cross-view kernel. As the definition of the single-view kernels, we take the form of Gaussian function to define the cross-view kernel function:
\begin{equation}
\bm{K}_{xy}(\bm{x}_i,\bm{y}_j) = exp(-d_{xy}^2(\bm{x}_i,\bm{y}_j)/2\sigma_{xy}^2)
\label{Eq128}
\end{equation}

The most important component in such kernel function is $d_{xy}(\bm{x}_i,\bm{y}_j)$, which defines a distance between one pair of Euclidean and Riemannian data on the underlying Euclidean space and Riemannian manifold. According to the type of the involved Riemannian manifold, it is expected to define the cross-view metric in three different heterogeneous matching/fusing cases, i.e., Euclidean-to-Grassmannian (EG), Euclidean-to-AffineGrassmannian (EA) and Euclidean-to-SPD (ES) cases, in the following.

\textbf{For Eucliean-to-Grassmannian case}:

In Nearest Feature Subspace (NFS) classifier \cite{chien2002discriminant}, the sets are modeled as linear subspaces, which reside on Grassmann manifold. The NFS calculates the distance between Euclidean point $\bm{x}_i$ and the linear subspace representation $\bm{U}_j$ in the following:
\begin{equation}
d_{xy}(\bm{x}_i,\bm{y}_j) = \|\bm{x}_i-\bm{U}_j\bm{U}_j^T\bm{x}_i\|_\mathcal{F}.
\label{Eq4}
\end{equation}

\textbf{For Eucliean-to-AffineGrassmannian case}:

K-local Hyperplane Distance Nearest Neighbor (HKNN) algorithm \cite{vincent2001k} models the sets as affine subspaces lying on affine Grassmann manifold. It defines the distance between the Euclidean point $\bm{x}_i$ and the affine subspace $\bm{A}_j$ as:
\begin{equation}
d_{xy}(\bm{x}_i,\bm{y}_j) = \min_{\alpha} \|(\bm{U}_j\alpha+\bm{\mu}_j)-\bm{x}_i\|_2.
\label{Eq6}
\end{equation}
where $\alpha$ is a vector of coordinates for the points within $\bm{A}_j$.


\begin{figure}[t]
	\begin{center}
		\includegraphics[width=0.9\linewidth]{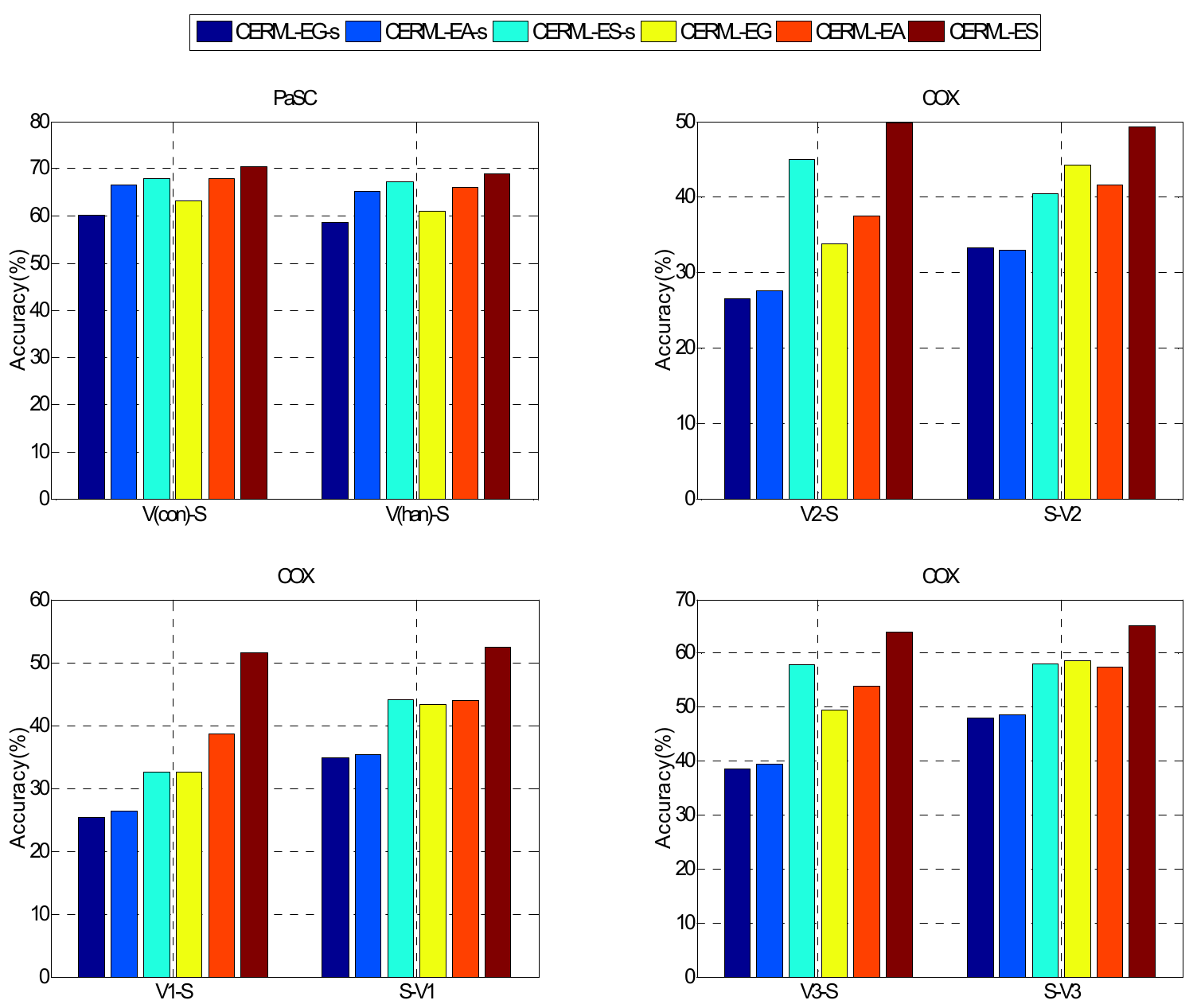}
	\end{center}
	\caption{V2S/S2V face recognition results (\%) of the proposed CEMRL dealing with different representations of videos on PaSC (deep feature) and COX (gray feature). Here, CERML-EG, CERML-EA, CERML-ES respectively indicates videos are represented by mean+subsapce, mean+affine subspace, mean+SPD matrix. CERML-EG-s, CERML-EA-s, CERML-ES-s are the cases with single-view kernel while the other three cases are those adding cross-view kernels.}
	\label{Fig1}
\end{figure}

\begin{table*}[t]
	\linespread{1.2}
	\caption{Running time (seconds) in the task of V2S/S2V face recognition on COX. Here EG, EA and ES indicate the Euclidean-to-Grassmannian, Euclidean-to-AffineGrassmannian and Euclidean-to-SPD matchings.}
	\begin{center}
		\footnotesize
		\begin{tabular}{|c|ccccccccc|}
			\hline
			Methods & NCA & ITML & LFDA & LMNN & PSDML & KPLS-EG & KPLS-EA & KPLS-ES & KCCA-EG \\
			\hline\hline
			
			Train  & 10040.59 & 2520.41 & 79.28 & 470.11 & 1350.55 & 644.29 & 2547.62 & 468.94 & 639.51  \\
			Test    & 0.14 & 1.57 & 0.14 & 0.14  & 0.34 & 0.94 & 3.24 & 0.63 & 0.94  \\
			\hline
			Methods & KCCA-EA & KCCA-ES & KGMA-EG & KGMA-EA & KGMA-ES & CERML-EG & CERML-EA & CERML-ES &   \\
			\hline\hline
			
			Train  & 2542.95 & 464.16 & 645.42 & 2548.83  & 470.07 & 754.13 & 2657 & 578.86 &   \\
			Test    & 3.24 & 0.63 & 0.94  & 3.24  & 0.63 & 0.94 & 3.24 & 0.63 &   \\
			\hline
		\end{tabular}
	\end{center}
	\label{tab_time_v2s}
\end{table*}

\begin{table*}[t]
	\linespread{1.2}
	\caption{Running time (seconds) in the task of V2V face recognition on COX. Here EG, EA and ES represent the Euclidean-to-Grassmannian, Euclidean-to-AffineGrassmannian and Euclidean-to-SPD matchings.}
	\begin{center}
		\footnotesize
		\begin{tabular}{|c|ccccccccccc|}
			\hline
			Methods  & DCC & GDA & GGDA &  AHISD & CHISD & SSDML & CDL & LMKML  & CERML-EG & CERML-EA & CERML-ES \\
			\hline\hline
			
			Train   & 189.79 & 118.06 & 168.67   & N/A & N/A & 7012.80 & 431.79 & 92968.91 & 260.94 & 944.72  & 251.07\\
			Test    & 8.79 & 0.29 & 0.36  & 8.85 & 27.46 & 3.23 & 5.77 & 1.07 & 0.21 & 1.53  & 0.20\\
			
			\hline
		\end{tabular}
	\end{center}
	\label{tab_time_v2v}
\end{table*}

\textbf{For Euclidean-to-SPD case}:

Classical Mahalanobis Distance (MD) can be used to define the distance between the Euclidean point $\bm{x}_i$ and the covariance matrix $\bm{C}_j$, which is commonly treated a SPD matrix and thus residing on SPD manifold:
\begin{equation}
d_{xy}(\bm{x}_i,\bm{y}_j) = \sqrt{(\bm{x}_i-\bm{\mu}_j)^T\bm{C}_j^{-1}(\bm{x}_i-\bm{\mu}_j)}.
\label{Eq77}
\end{equation}
where  $\bm{\mu}_j$ is the mean of the samples in the set.


To validate the benefit of employing cross-view kernels aforementioned for the proposed CERML, as shown in Fig.\ref{Fig1}, we present the detailed results of the proposed CERML methods with or without using the cross-view kernels in the evaluations of V2S/S2V face recognition on PaSC \cite{beveridge2013challenge} and COX \cite{huang2015cox}.

From these comparisons, we can draw the same conclusion that our proposed CERML adding cross-view kernel typically perform better than that only using single-view kernel, demonstrating the superiority of our CERML with cross-view kernel working in the task of V2S/S2V face recognition.

\section{Running time of the evaluated methods}

In this section, we tabulates the training and testing time of the competing methods working in the three video-based face recognition scenarios on an Intel(R) Core(TM) i7-3770M (3.40GHz) PC. Note that the running time of the comparing  methods includes the computation of the involved Riemannian representations such as SPD representations.

In the evaluation of V2S/S2V face recognition, we compare the following state-of-the-art Euclidean/Riemannian metric learning methods:
\begin{enumerate}
	%
	
	\item Homogeneous (Euclidean) metric learning methods:
	
	Neighbourhood Components Analysis (NCA) \cite{goldberger2004neighbourhood},  Information-Theoretic Metric Learning
	(ITML)\cite{davis2007information}, Local Fisher Discriminant Analysis (LFDA) \cite{sugiyama2007dimensionality}, Large Margin Nearest Neighbor (LMNN) \cite{weinberger2009distance} and Point-to-Set Distance Metric Learning \mbox{(PSDML) \cite{zhu2013p2s}};
	
	\item Heterogeneous metric learning methods:
	
	Kernel Partial Least Squares (KPLS) \cite{sharma2011pls}, Kernel Canonical Correlation Analysis (KCCA) \cite{hardoon2004canonical} and Kernel Generalized
	Multiview Linear Discriminant Analysis (KGMA) \cite{sharma2012gma}.
	
\end{enumerate}

In Tab.\ref{tab_time_v2s}, the running times (seconds) of the comparative methods in the task of V2S/S2V face recognition on COX are presented. In this table, training time is reported to study the running speed of the evaluated method working on the training data. For testing, we report the average classification time for recognizing 1 probe subject from 700 gallery subjects. As seen, the two Euclidean metric learning methods NCA and ITML and the four Euclidean-to-Riemannian metric approaches KPLS-EA, KCCA-EA, KGMA-EA and CERML-EA require comparatively more time for training. Excluding these techniques, the running time of other methods including our CERML are comparable.

In the evaluation of V2V face recognition, we compare the following state-of-the-art Riemannian metric learning methods:

\begin{enumerate}
	\item Grassmannian metric learning methods:
	
	Discriminative Canonical Correlations (DCC) \cite{kim2007discriminative}, Grassmann Discriminant Analysis (GDA) \cite{hamm2008gda}, Grassmannian Graph-Embedding Discriminant Analysis (GGDA) \cite{harandi2011ggda};
	
	\item Affine Grassmannian metric learning methods:
	
	Affine Hull based Image Set Distance (AHISD) \cite{cevikalp2010face}, Convex Hull based Image Set Distance (CHISD) \cite{cevikalp2010face}, Set-to-Set Distance Metric Learning (SSDML) \cite{zhu2013p2s};
	
	\item SPD Riemannian metric learning methods:
	
	Localized Multi-Kernel Metric Learning (LMKML) \cite{lu2013image}, Covariance Discriminative Learning (CDL) \cite{wang2012covariance}.
	
\end{enumerate}

In Tab.\ref{tab_time_v2v}, the running times (seconds) of the competing methods in the scenario of V2V face recognition on COX are presented. In this table, training time is only required by supervised methods. For testing, we report the classification time for recognizing 1 probe subject from 700 gallery subjects. As seen, the Riemannian metric learning methods SSDML, LMKML and CERML-EA need much more time than other competing methods in training, while DCC, AHISD and CHISD runs much lower than others in testing. Besides to them, the running times of other methods including our CERML are comparable working on COX.

\end{document}